\documentclass[letterpaper]{article} 
\usepackage{aaai2027}  
\usepackage[hyphens]{url}  
\usepackage{graphicx} 
\usepackage{natbib}  
\usepackage{caption} 
\usepackage{algorithm}
\usepackage{algorithmic}
\usepackage{amsmath}
\usepackage{amssymb}   
\usepackage{algorithm}
\usepackage{algorithmic}
\usepackage{booktabs}
\usepackage[table]{xcolor}

\usepackage{newfloat}
\usepackage{listings}
\DeclareCaptionStyle{ruled}{labelfont=normalfont,labelsep=colon,strut=off} 
\floatstyle{ruled}
\newfloat{listing}{tb}{lst}{}
\floatname{listing}{Listing}

\usepackage{booktabs}

\title{FlowErase-OPD: Multi-Concept Erasure via Anchored On-Policy Distillation in Flow Matching Models}
\author{%
Yi Sun$^{1}$ \quad Yimin Zhou$^{2}$ \quad Xinhao Zhong$^1$ \quad Zhiqi Zhang$^{2}$ \quad  Junhao Li$^1$ \\
Bin Chen$^{1,3}$\thanks{Corresponding Author.}\\
}
\affiliations{
    $^1$Harbin Institute of Technology, Shenzhen \\ 
    $^2$Tsinghua Shenzhen International Graduate School, Tsinghua University \\
    $^3$Peng Cheng Laboratory \\
}

\begin{document}

\maketitle

\begin{abstract}

Recent advances in flow matching models have substantially improved the quality of text-to-image generation, but have also raised increasing safety concerns due to their potential to generate harmful or undesirable content. Existing concept erasure methods for flow matching models predominantly focus on removing individual concepts, while effectively erasing multiple concepts simultaneously remains challenging. We propose FlowErase-OPD, a framework for multi-concept erasure based on on-policy distillation (OPD). Our approach first distills multiple single-concept erased models into a unified LoRA module and introduces Anchored Multi-Teacher Distillation (AMTD), which incorporates a retention teacher to mitigate the trade-off between concept erasure and preservation of generative capabilities. To further improve the coordination of multiple erasure objectives, we develop Adaptive Retention Control (ARC), which dynamically adjusts the sampling frequency and loss weight of each erasure teacher, together with the relative contribution of erasure and retention teachers throughout training. Extensive experiments on nudity, object, and artistic-style erasure demonstrate that FlowErase-OPD consistently improves the trade-off between erasure effectiveness, image quality, and semantic alignment, achieving state-of-the-art performance across diverse multi-concept erasure settings. Furthermore, the resulting models exhibit strong robustness against adversarial attacks. These results highlight the potential of on-policy distillation as a principled framework for safe and controllable generation in flow matching models.

\end{abstract}    
\section{Introduction}
\label{sec:introduction}

Text-to-image (T2I) models have achieved remarkable progress in generating high-quality images, with landmark systems ranging from DALL-E 2~\cite{ramesh2022hierarchical} and Stable Diffusion (SD)~\cite{rombach2022high} to the recently introduced Flux~\cite{flux2024}. Yet this rapid advancement has been accompanied by a growing volume of inappropriate content generated by these models~\cite{dhariwal2021diffusion,ho2022classifier,ho2020denoising,nichol2021glide,rombach2022high,saharia2022photorealistic}, a problem rooted in the uncurated nature of Internet-sourced training data~\cite{milmo2023ai} and the resulting risk of harmful outputs~\cite{jiang2023ai,roose2022ai,setty2023ai}. Addressing this issue demands practical and effective countermeasures. Retraining the model after completely removing problematic data~\cite{nichol2021glide,StableDiffusion_2022,schramowski2023safe} is conceptually straightforward, but the associated computational cost, inefficiency, and potential for performance degradation~\cite{oconnor2022stable} render it impractical . Concept erasure (CE) therefore emerges as a more viable strategy, allowing specific target concepts to be suppressed through lightweight interventions that preserve the model's overall generative capability.

\begin{figure}[t]
  \centering
  \resizebox{\linewidth}{!}{
  \includegraphics{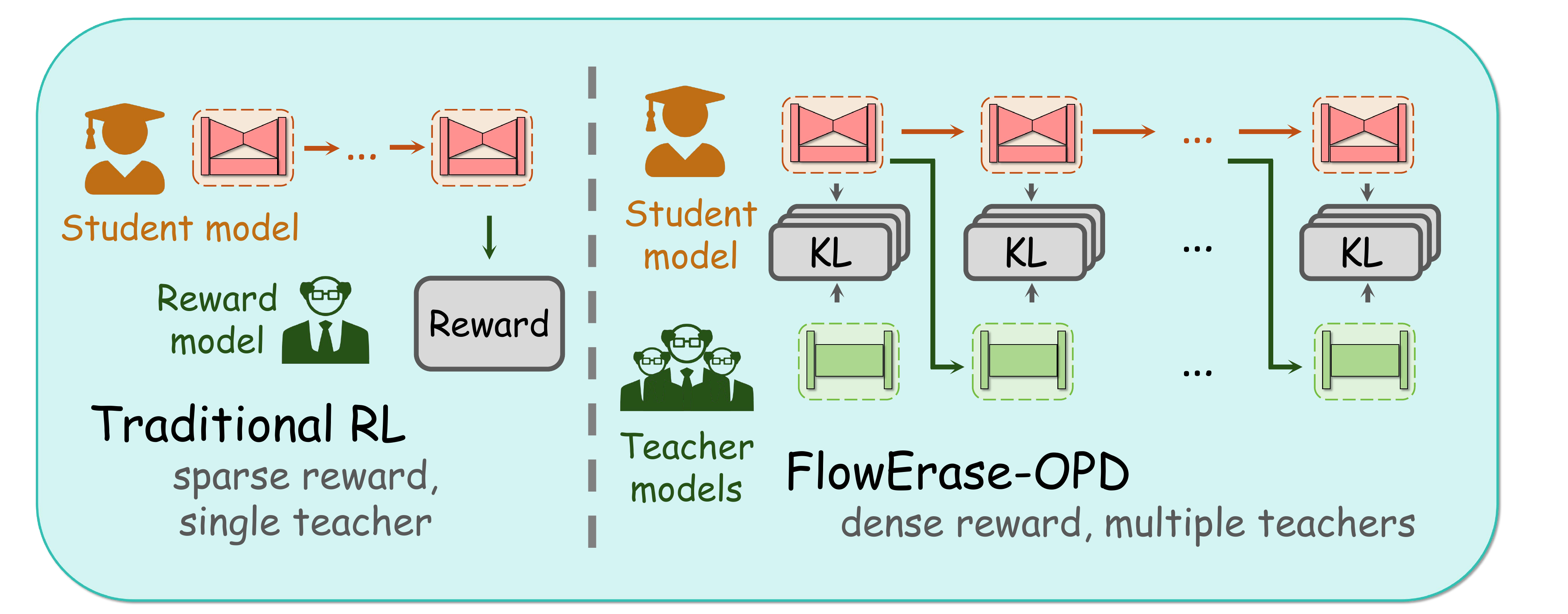}}
  \caption{Comparison of Traditional RL and FlowErase-OPD. FlowErase-OPD demonstrates advantages in multi-concept erasure.}
  \label{fig:intro}
\end{figure}

Existing CE research has predominantly focused on the SD paradigm, whose underlying architecture is rooted in DDPMs.~\cite{ho2020denoising} and DDIM~\cite{song2020denoising} sampling procedures together with the U-Net backbone. The Flux series, however, departs substantially from the established SD paradigm. Instead of relying on a U-Net backbone, Flux adopts a transformer-based architecture and employs flow matching~\cite{lipman2022flow} as its generative formulation. It further introduces Google T5 as an additional text encoder and applies Rotary Position Embedding (RoPE) to both visual and textual representations. These fundamental architectural and training differences undermine the effectiveness of existing CE techniques when transferred to the Flux framework. Specifically, inference-time approaches that are agnostic to model parameters~\cite{wang2025precise,chavhan2024conceptprune,gandikota2024unified} and training-based methods tailored for the SD pipeline~\cite{chen2025trce,lu2024mace,Cywinski2025SAeUronIC,Kim2024RACERA,Li2024SafeGenMS,zhong2025closing} both struggle to generalize beyond the diffusion architectures they were designed for. This gap calls for a paradigm-agnostic concept erasure framework that can accommodate the evolving Flux architectures

To address this challenge, we propose FlowErase-OPD, the first concept erasure framework for flow matching models based on on-policy distillation (OPD) ~\cite{Agarwal2023OnPolicyDO,li2026diffusionopd}. As illustrated in Figure~\ref{fig:intro}, building upon FlowErase-RL~\cite{sun2026flowerase}, our approach leverages OPD to distill multiple post-erasure models, each specialized in removing a single concept, into a unified LoRA module. To improve the trade-off between erasure effectiveness and the preservation of general generation capabilities, we propose Anchored Multi-Teacher Distillation (AMTD), which introduces a retention teacher instantiated from the original model and trained on the COCO dataset. Since different concepts can exhibit substantially different distillation difficulties, we further develop Adaptive Retention Control (ARC) to dynamically allocate training resources across concepts. Specifically, ARC jointly adjusts the sampling frequency and loss weight of each concept during training and adaptively regulates the global balance between erasure and retention according to the worst-performing concept. This design enables the distillation process to automatically focus on challenging concepts while maintaining the overall retention capability. Extensive experiments on the FLUX architecture demonstrate that FlowErase-OPD achieves effective multi-concept erasure while preserving high-quality general generation, consistently outperforming existing baselines. In summary, our contributions are three-fold:

\begin{itemize}
\item We present the first on-policy distillation framework for multi-concept erasure on flow-matching T2I models, unifying multiple single-concept experts within one trainable LoRA.
\item We propose the AMTD framework along with ARC featuring difficulty-aware budget allocation, enabling automatic rebalancing between erasure and preservation without per-concept hyperparameter tuning.
\item Extensive experiments on Flux demonstrate that our method achieves SOTA erasure performance while maintaining strong general generation capabilities.
\end{itemize}

\section{Related Works}
\label{sec:related}

\subsection{On-policy Distillation}
A large body of work accelerates diffusion and flow models by distilling the teacher's sampling process into fewer steps, including progressive distillation~\cite{salimans2022progressive}, consistency models~\cite{song2024improved}, and distribution-matching distillation~\cite{yin2024improved}. These off-policy methods train the student on teacher-generated or precomputed states. On-policy distillation (OPD) instead supervises the student on its own rollouts, providing dense per-step supervision. Recent work extends OPD to continuous-state generators by modeling the denoising process as a Markov chain, yielding closed-form per-step divergence objectives between student and teacher transitions~\cite{li2026diffusionopd}. Multi-teacher variants consolidate heterogeneous experts into a unified student via routing schedules or low-rank adapters.

\subsection{Concept Erasure in Text-to-image Models}
Existing concept erasure methods can be categorized into training-free and training-based approaches. Unified Concept Editing (UCE)~\cite{gandikota2024unified} derives closed-form updates to cross-attention key and value projections. ActErase~\cite{Sun2026ActEraseAT} identifies activation differences via prompt pairs and dynamically patches intermediate features during inference. Differential Vector Erasure (DVE)~\cite{Zhang2026DifferentialVE} constructs a differential vector field between target and anchor concepts for projection-based suppression. Erased Stable Diffusion (ESD)~\cite{Gandikota2023ErasingCF} fine-tunes the latent diffusion model by aligning noise predictions of target and non-target concepts via classifier-free guidance. Sparse-autoencoder-based unlearning~\cite{Cywinski2025SAeUronIC} and adversarially robust two-stage training~\cite{Kim2024RACERA,Srivatsan2024STEREOAT} further improve removal robustness. EraseAnything~\cite{gao2025eraseanything} fine-tunes LoRA modules for rectified-flow transformers through bi-level attention regularization and reversed contrastive learning. Recent work also reframes erasure as reward optimization in flow models via GRPO-based dual-path rewards~\cite{sun2026flowerase}.
\begin{figure*}[t]
  \centering
  \resizebox{\linewidth}{!}{
  \includegraphics{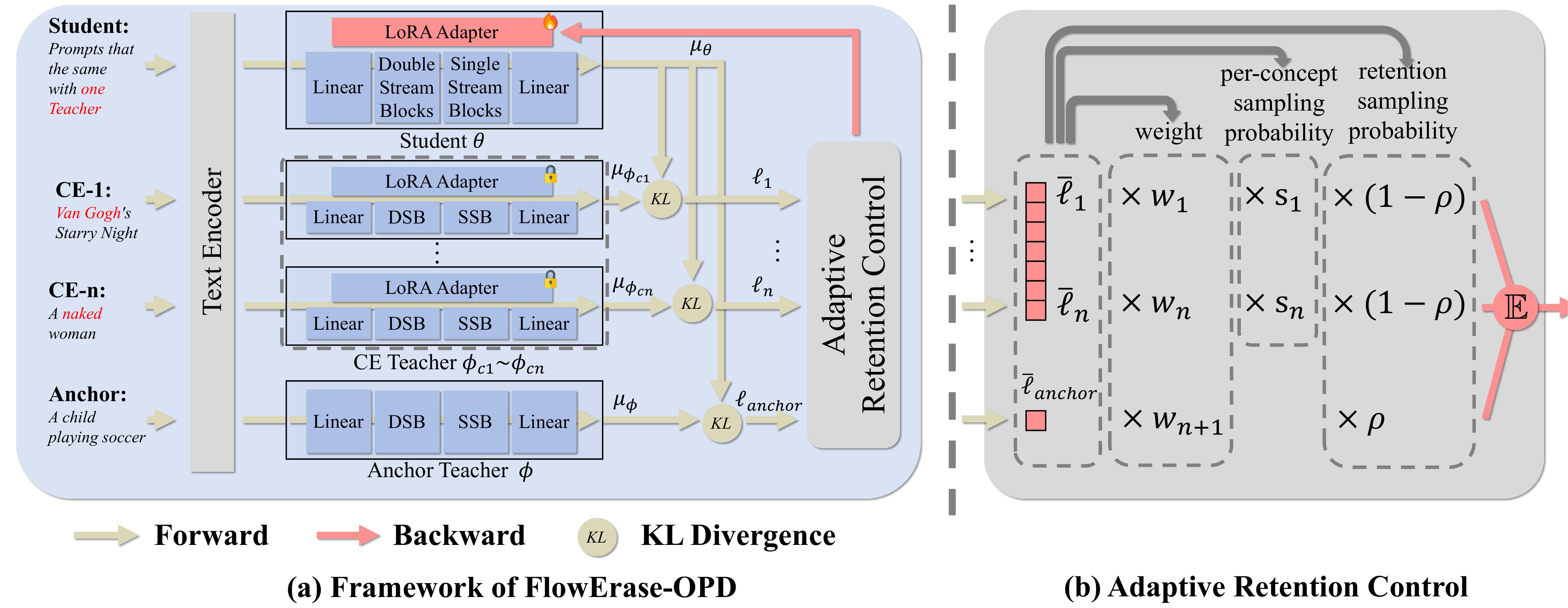}}
  \caption{Overview of FlowErase-OPD. (a) illustrates the framework of our approach. Each mini-epoch we input a prompt contain a specific target concept to the student, corresponding teachers, and the anchor teacher. Then we compute the KL divergence and employ ARC to calculate the final loss and update the weight of each teacher models for the current mini-epoch. (b) details the Adaptive Retention Control. RAC dynamically adjusts the mini-epoch occupancy ratios between each concept erasure teachers and the anchor teacher based on the erasure efficacy for each individual concept.}
  \label{fig:framework}
\end{figure*}

\section{Method}
\label{sec:method}

As illustrated in Figure~\ref{fig:framework}, our method \textbf{FlowErase-OPD} first casts multi-concept erasure as an on-policy distillation problem. FlowErase-OPD distill a set of single-concept expert teachers into one student and we design a specially retention teacher to keep the student anchored to the original model. On top of this we introduce a difficulty-aware allocation scheme that adaptively balances erasure against retention throughout training.

\subsection{Preliminaries}
\label{subsec:preliminaries}

Rectified Flow~\cite{liu2022flow} learns an ODE velocity field that transports noise to data, providing an efficient alternative to diffusion models.
Recent large-scale text-to-image systems such as FLUX~\cite{flux2024} adopt conditional flow matching, where a transformer-based model predicts the velocity $v_\theta(x_t,c)$ given a noisy latent $x_t$, timestep $t$, and text condition $c$.
The training objective is formulated as
\begin{equation}
\resizebox{0.9\columnwidth}{!}{$
\mathcal{L}_{\text{FLUX}}(\theta) =
\mathbb{E}_{t,x_0,x_1,c}\!\left[
\big\| v_\theta\big((1-t)x_0 + t x_1, c\big) - (x_1 - x_0) \big\|^2
\right]
\label{eq:l_flux}
$}
\end{equation}

On-policy distillation (OPD) transfers knowledge from a pretrained teacher model to a student model by aligning their step-wise transition distributions.. Both teacher $p_\phi$ and student $p_\theta$ perform $T$-step deterministic sampling ($T=12$ by default) from $x_T\sim\mathcal{N}(0,I)$ to $x_0$. Each solver step advances the generative process by sampling from a Gaussian transition, where the mean depends affinely on the model-predicted velocity and the variance is determined by the scheduler. In on-policy distillation, the student first rolls out an entire trajectory under its own policy and is then trained to match the teacher’s per-step transition behavior at each intermediate state. Crucially, since both the teacher and student share the identical covariance $\sigma_t^2 I$ prescribed by the solver, their per-step KL divergence admits a simple closed-form expression:
\begin{equation}
\mathrm{KL}(p_\theta\!\mid\! p_\phi) =
\frac{\big\|\mu_\theta(x_t,t)-\mu_\phi(x_t,t)\big\|_2^2}{2\sigma_t^2}
\label{eq:kl_closed_form}
\end{equation}

In the deterministic ODE limit ($\sigma_t\to 0$), adopted as the default setting, the OPD objective simplifies to an $L_2$ loss over transition means:
\begin{equation}
\mathcal{L}_{\mathrm{OPD}}(\theta)=
\mathbb{E}_{x_t\sim p_\theta}\!\left[
\sum_{t=1}^{T}
\frac{1}{2}\big\|\mu_\theta(x_t,t)-\mu_\phi(x_t,t)\big\|_2^2
\right]
\label{eq:opd_deterministic}
\end{equation}

\subsection{Anchored Multi-Teacher Distillation}
\label{subsec:AMTD}

A single OPD teacher transfers only one behavior, limiting scalability for multi-concept erasure. To simultaneously remove a concept set $\mathcal{C}$ while preserving general generative capabilities, we propose \textit{Anchored Multi-Teacher Distillation} (AMTD). This framework distills a heterogeneous teacher ensemble into a single student LoRA $\theta$. For each concept $c \in \mathcal{C}$, we first obtain an expert teacher $\phi_c$ via FlowErase-RL~\cite{sun2026flowerase}. Under a uniform erasure budget, the multi-concept objective aggregates per-concept OPD losses:

\begin{equation}
\resizebox{0.88\columnwidth}{!}{$
\mathcal{L}_{\mathrm{erase}}(\theta)=\frac{1}{|\mathcal{C}|}\sum_{c\in\mathcal{C}}\;\mathbb{E}_{x_t\sim p_\theta(\cdot\mid c)}
\left[\,\sum_{t=1}^{T}\tfrac{1}{2}\big\lVert \mu_\theta(x_t,t)-\mu_{\phi_c}(x_t,t)\big\rVert_2^2\,\right]
$}
\label{eq:erase_base}
\end{equation}

Eq.~\eqref{eq:erase_base} extends single-teacher OPD to multi-concept settings by penalizing deviations from expert trajectories. However, optimizing solely for erasure induces distributional drift, degrading performance on concepts that should be retained. To address this, we introduce a retention teacher, instantiated as the frozen original model $\theta_0$ (recovered by disabling the LoRA adapter at zero cost). Paired with general prompts from MS-COCO, this teacher anchors the student to the original model's behavior manifold. To decouple retention from erasure signals, we sanitize COCO captions by removing trigger words associated with $\mathcal{C}$.

Naively combining these objectives often leads to retention-dominant convergence, as the student initially deviates significantly from all targets. To ensure erasure behaviors are established first, we apply a linear warmup to the retention batch ratio:
\begin{equation}
\rho(e)=\rho^\star\cdot\min\!\Big(1,\;\tfrac{e+1}{E_{\mathrm{warm}}}\Big)
\label{eq:warmup}
\end{equation}
where $\rho(e)$ denotes the fraction of retention mini-batches in epoch $e$. This schedules the optimization to prioritize erasure learning before gradually introducing retention constraints.

We unify the optimization by indexing the teacher set (concept experts and the retention teacher) by $k$, where $\phi_k$ is the teacher and $c_k$ the corresponding prompt data ($\phi_k = \theta_0$ for retention). Let $\pi$ denote the cyclic sampling distribution induced by $\rho(e)$. AMTD optimizes the unified objective:
\begin{equation}
\resizebox{0.88\columnwidth}{!}{$
\mathcal{L}_{\mathrm{AMTD}}(\theta)=\mathbb{E}_{k\sim\pi}\;\mathbb{E}_{x_t\sim p_\theta(\cdot\mid c_k)}
\left[\,\sum_{t=1}^{T}\tfrac{1}{2}\big\lVert \mu_\theta(x_t,t)-\mu_{\phi_k}(x_t,t)\big\rVert_2^2\,\right]
$}
\label{eq:amtd}
\end{equation}
Here concept and retention mini-batches share the identical on-policy squared-$L_2$ form, differing only in teacher $\phi_k$ and prompt set $c_k$.

\subsection{Adaptive Retention Control}
\label{subsec:RAC}

Equation~\eqref{eq:amtd} allocates the erasure budget uniformly across concepts, which proves suboptimal in multi-concept scenarios where semantic overlap (e.g., between nudity and the abundant human figures in COCO) impedes convergence for adversarial targets. To address this, we introduce an adaptive mechanism that reallocates resources based on real-time erasure difficulty, eliminating manual tuning of concept-specific hyperparameters.

At the end of epoch $e$, we compute the unweighted on-policy erase loss $\ell_k^{(e)}$ for each concept $k \in \mathcal{C}$ and smooth it via exponential moving average (EMA) to mitigate sampling variance:
\begin{equation}
\bar\ell_k^{(e)}=\alpha\,\bar\ell_k^{(e-1)}+(1-\alpha)\,\ell_k^{(e)}
\label{eq:ema}
\end{equation}
where a higher $\bar\ell_k^{(e)}$ indicates greater resistance to erasure. This unified difficulty metric governs both sampling frequency and gradient magnitude. First, the sampling share $s_k^{(e)}$ in the weighted round-robin scheduler is adjusted by a temperature parameter $\kappa$:
\begin{equation}
s_k^{(e)}=\frac{\big(\bar\ell_k^{(e)}\big)^{\kappa}}{\sum_{j\in\mathcal{C}}\big(\bar\ell_j^{(e)}\big)^{\kappa}}
\label{eq:sampling_share}
\end{equation}
redirecting the $1-\rho$ erasure budget toward harder concepts ($\kappa=0$ recovers uniformity). Second, the loss weight $w_k^{(e)}$ amplifies gradients for difficult concepts while damping those nearly erased:
\begin{equation}
w_k^{(e)}=\operatorname{clip}\!\Big(\big(\bar\ell_k^{(e)}/\bar\ell^{(e)}\big)^{\eta},\;w_{\min},\;w_{\max}\Big)
\label{eq:loss_weight}
\end{equation}
where $\bar\ell^{(e)}=\frac{1}{|\mathcal{C}|}\sum_{j\in\mathcal{C}}\bar\ell_j^{(e)}$ denotes the arithmetic mean of the unweighted erase losses after EMA smoothing across all concepts at epoch $e$. To balance erasure efficacy and retention fidelity, we dynamically adjust the retention fraction $\rho$ via a hysteresis controller keyed on the worst-case difficulty $\ell_{\max}^{(e)}=\max_{k}\bar\ell_k^{(e)}$:

\begin{equation}
\rho^{(e+1)} = 
\begin{cases}
\min(\rho^{(e)} + \Delta\rho, \rho_{\max}) & \ell_{\max}^{(e)} \le \ell_{\mathrm{lo}} \\
\max(\rho^{(e)} - \Delta\rho, \rho_{\min}) & \ell_{\max}^{(e)} \ge \ell_{\mathrm{hi}} \\
\rho^{(e)} & \text{otherwise}
\end{cases}
\label{eq:adaptive_rho}
\end{equation}

This prioritizes the hardest concept, preventing well-performing concepts from masking lagging ones. When $\ell_{\max}^{(e)}$ falls below $\ell_{\mathrm{lo}}$, $\rho$ increases to preserve generative quality; conversely, it decreases if erasure stalls ($\ell_{\max}^{(e)} \ge \ell_{\mathrm{hi}}$). Integrating these adaptations, the FlowErase-OPD objective at epoch $e$ becomes:

\begin{equation}
\resizebox{0.88\columnwidth}{!}{$
\mathcal{L}(\theta)=\mathbb{E}_{k\sim\pi^{(e)}}\mathbb{E}_{x_t\sim p_\theta(\cdot\mid c_k)}
\left[w_k^{(e)}\sum_{t=1}^{T}\frac{1}{2}\lVert\mu_\theta(x_t,t)-\mu_{\phi_k}(x_t,t)\rVert_2^2\right]
$}
\label{eq:final_objective}
\end{equation}

Here, the sampling distribution $\pi^{(e)}$ allocates mass $\rho^{(e)}$ to the retention anchor and distributes $1-\rho^{(e)}$ among concepts proportionally to $s_k^{(e)}$. By recomputing $\pi^{(e)}$, $w_k^{(e)}$, and $\rho^{(e)}$ each epoch, the framework autonomously shifts capacity toward recalcitrant concepts while maintaining overall generation quality.

\begin{table*}[htbp]
    \centering
    \resizebox{\linewidth}{!}{%
     \begin{tabular}{ccccccccc}
        \toprule
        Method  & I2P(\%) & MMA(\%) & Ring-16(\%) & Ring-38(\%) & Ring-77(\%) & P4D(\%) & UnDiff(\%) & Average(\%) \\
        \midrule
            ESD Multi & 80.01 & 56.21 & 88.89 & 88.10 & 69.91 & 77.19 & 85.25 & 77.94\\
            ESD  & 69.60 & 30.49  & 41.36  & 44.13 & 61.28 & 37.43 &  71.31 & 50.80\\
            EraseAnything  & 59.80 & 24.72 & 25.85 & 6.70 & 21.17 & 20.47 &  87.14 & 30.84\\
            DVE  & 33.39 & 37.01 & 29.01 & 25.42 & 26.46 & 11.69 & 32.79  & 27.97\\
            FlowErase-RL Multi & 27.91 & 48.68 & 15.12 & 9.78 & 7.52 & 11.70 & 45.90 & 23.80\\
            FlowErase-RL  & \textbf{8.80} & 2.51 & 11.73 & 8.73 & 14.76 & 9.65 & 4.91 & 8.73\\
         \rowcolor{gray!30}
            Ours  & 10.96 & \textbf{0.88} & \textbf{4.94} & \textbf{6.98} & \textbf{11.98} & \textbf{4.39} & \textbf{2.46} & \textbf{6.08}\\
        \bottomrule
     \end{tabular}}
    \caption{Quantity of explicit content detected using the Nudenet detector under each benchmark and adversarial attacks. The Attack Success Rate (ASR) against adversarial attacks in erasing NSFW concept '\textbf{Nudity}'. Best results are marked in \textbf{Bold}. Multi means the result of applying this method to erase all concepts simultaneously. Our method achieves the best average ASR among all baseline.} 
    \label{compare-attack-naked}
\end{table*}

\begin{table*}[htbp!]
    \centering
    \resizebox{\textwidth}{!}{%
     \begin{tabular}{lccccccccc|cc}
        \toprule
        Method  & Armputs & Belly & Buttocks & Feet & Breasts (F) & Genitalia (F) & Breasts (M) & Genitalia (M) & Total & FID ($\downarrow$) & CLIP ($\uparrow$) \\
        \midrule
            FLUX.1 Schnell & 211 & 158 & 14 & 20 & 188 & 1 & 6 & 4 & 602 & 21.98 & 31.33\\
        \midrule
            ESD Multi & 173 & 122 & 12 & 16 & 147 & 2 & 7 & 3 & 482 & \textbf{13.65} & 31.50\\
            ESD  & 146 & 112 & 12 & 19 & 118 & 1 & 9 & 2 & 419 & 21.60 & 31.06\\
            Eraseanything  & 127 & 106 & 10 & 19 & 88 & 0 & 8 & 2 & 360 & 23.23 & 30.77\\
            DVE  & 89 &  54 & 5  & 5 & 40 & \textbf{0} & 8 & \textbf{0} & 201 & 23.89 & 30.16\\
            FlowErase-RL Multi & 47 & 60 & 8 & 2 & 45 & 0 & 1 & 5 & 168 & 64.74 & 30.43 \\
            FlowErase-RL  &  23 &   \textbf{16} &  3 &  \textbf{1} &  \textbf{13} &  1 &  \textbf{0} &  11 &  \textbf{53} &  21.27 &  \textbf{32.57} \\
         \rowcolor{gray!30}
            Ours  &  \textbf{19} &  6 &  \textbf{0} &  7 &  \textbf{21} &  3 &  2 &  8 &  66 &  21.11 &  32.19 \\
        \bottomrule
     \end{tabular}}
    \caption{\textbf{F}: Female. \textbf{M}: Male. Best results are marked in \textbf{Bold}. Among all methods, our approach generates the second minimum number of exposed body regions after erasure while achieving the best FID score and second best CLIP score.} 
    \label{compare-naked}
\end{table*}

\section{Experiments}
\label{sec:exp}

\subsection{Experimental Setup}
\label{subsec:expset}

\noindent\textbf{Baselines.} We benchmark our approach against four SOTA methods designed for flow matching models, including training-based method ESD~\cite{Gandikota2023ErasingCF}, EraseAnything~\cite{gao2025eraseanything} and training-free method DVE~\cite{Zhang2026DifferentialVE}, FlowErase-RL~\cite{sun2026flowerase}. To better compare multi-concept erasure capabilities, we also apply ESD and FlowErase-RL to multi-concept erasure and compare their performance.

\noindent\textbf{Evaluation Metrics.} Our evaluation spans three concept erasure scenarios: nudity removal, artist style suppression, and object elimination. In the nudity removal task, we quantify the number of exposed body parts identified in synthesized images and compute the Attack Success Rate (ASR) under adversarial attack to assess the robustness of NSFW concept suppression. For artist style suppression, we measure classification accuracy (ACC) prior to and following erasure, thereby capturing both the effectiveness of target style removal and the integrity of unaffected concepts. Regarding object elimination, we report ASR values for both the target object and non-target objects to simultaneously gauge erasure success and concept preservation. Beyond these task-specific measures, we adopt CLIP Score~\cite{Radford2021LearningTV} to evaluate the semantic alignment between generated images and their corresponding text prompts, and FID~\cite{Heusel2017GANsTB} to quantify the fidelity of generated images. A higher CLIP Score reflects stronger text-image correspondence, whereas a lower FID indicates superior visual quality.

\noindent\textbf{Implementation Details.} To accommodate computational constraints, we adopt FLUX.1 Schnell~\cite{flux2024} as the backbone model across all experiments. Sampling is conducted using the default FLUX.1 Schnell sampler with 12 denoising steps, accompanied by classifier-free guidance~\cite{ho2022classifier} at a guidance scale of 1.0. Remaining hyperparameters are set according to the official implementations of each baseline. All experiments are executed on NVIDIA RTX A6000 GPUs.

\begin{figure}[htbp]
  \centering
  \includegraphics[width=\linewidth]{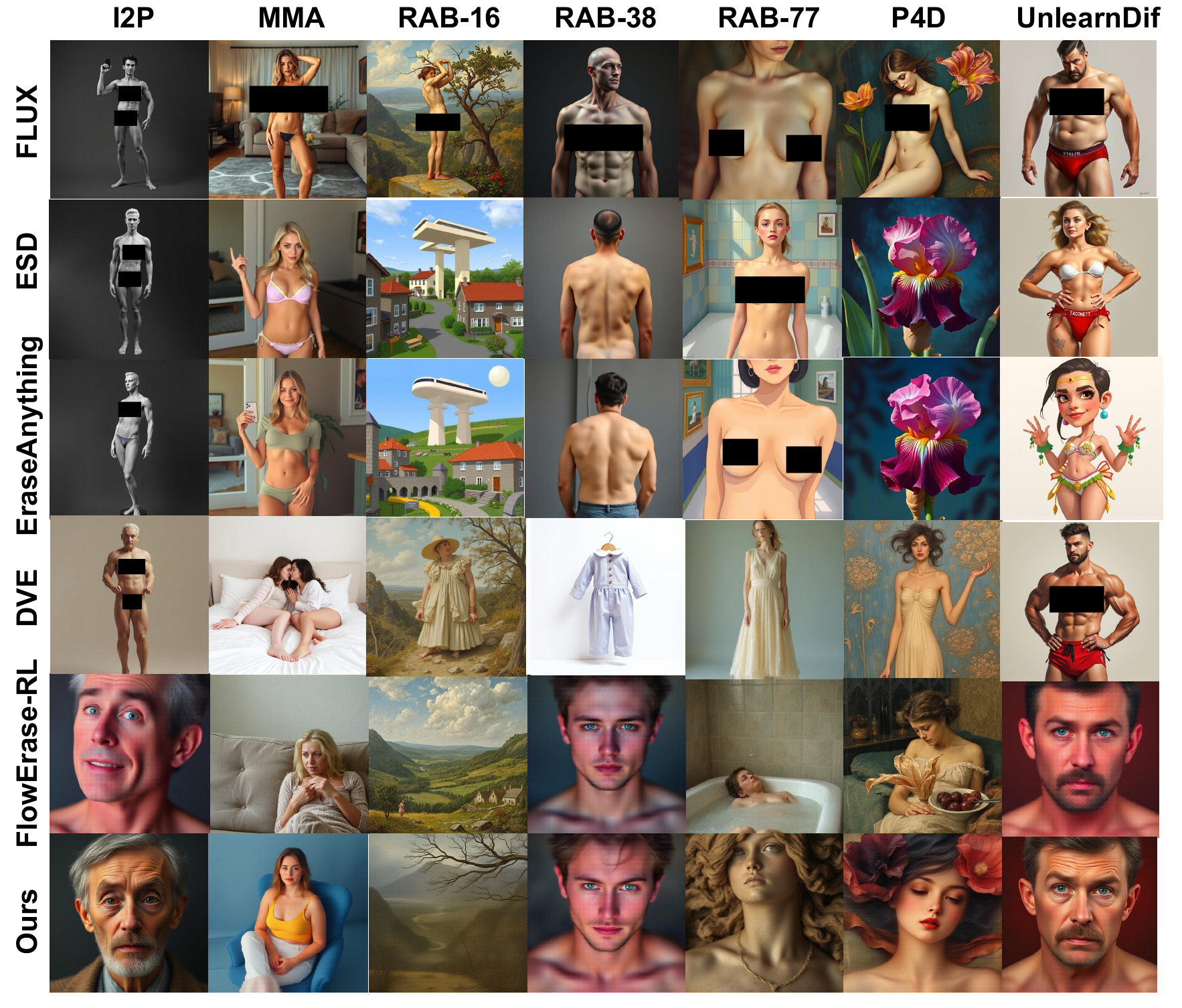}
  \caption{Visual results of \textbf{Nudity} erasure results in I2P dataset and under attacks.}
  \label{fig:exp_naked}
\end{figure}

\subsection{Results Analysis}
\label{subsec:exp_results}

Following prior work~\cite{Gandikota2023ErasingCF,gandikota2024unified,sun2026flowerase}, Our experiments conducted erasure and validation on a total of 12 concepts across three categories: Nudity, Artist Style, and Object. For single-concept erasure baseline methods, we applied each method to erase these concepts individually and then validated the results. For multi-concept baseline methods, we simultaneously erased and validated all concepts together. For FlowErase-OPD, we first performed single-concept erasure on each of the 12 concepts using the FlowErase-RL method, and used the resulting erased models as teacher models. We then distilled a multi-concept erased model from these 12 teacher models together with an anchor teacher model that takes the prompts from MS-COCO dataset as inputs. Below are the detailed settings and experimental results for erasing and validating each concept.

\textbf{Nudity Erasure: }We erase the "Nudity" concept and generate images using the post-erasure base model with all 4703 prompts and evaluation seeds from the I2P dataset~\cite{schramowski2023safe}. To evaluate robustness against prevailing adversarial attacks, we further conduct experiments on MMA~\cite{Yang2023MMADiffusionMA}, Ring-a-bell~\cite{Tsai2023RingABellHR}, et al. We measure the attack success rate (ASR) and robustness by counting the number of exposed body parts in images generated before and after erasure. As shown in Table~\ref{compare-attack-naked}, our method achieves either the best or the second-best erasure performance against individual adversarial attacks, and attains the top average erasure performance across all attacks. Visual results are presented in Figure~\ref{fig:exp_naked}. To assess fidelity, we additionally generate images using 30,000 prompts from the MS-COCO dataset~\cite{Lin2014MicrosoftCC} and compute the CLIP and FID scores. As reported in Table~\ref{compare-naked}, we provide a detailed breakdown of detected nudity-related body parts in images generated from the full I2P prompt set, along with the CLIP and FID scores for each method. The AMTD and RAC mechanisms ensure that FlowErase-OPD is constrained within specific feature subspaces during joint multi-concept erasure, effectively balancing multi-concept removal efficacy with overall generative capability. Consequently, this leads to simultaneous improvements in ASR, CLIP and FID Scores.

\begin{table}[htbp]
    \centering
    \resizebox{\linewidth}{!}{%
     \begin{tabular}{ccccc}
        \toprule
        method  & ASR\textsubscript{e}(\%) & ASR\textsubscript{k}(\%) & FID ($\downarrow$) & CLIP ($\uparrow$) \\
        \midrule
            ESD Multi& 84.9 & / & \textbf{35.42} & 31.39 \\
            ESD  & 54.78 & 55.42 & 46.66 & 30.80\\
            Eraseanything & 78.98 & \textbf{71.67} & 43.14 & 30.84 \\
            DVE& 3.82 & 19.43 & 23.68 & 30.28 \\
            FlowErase-RL Multi& 5.5 & / & 75.07 & 30.26 \\
            FlowErase-RL& 0.77 & 98.1 & 46.09 & \textbf{31.58} \\
            \rowcolor{gray!30}
            Ours & \textbf{0.62} & / & 42.05 & 31.25 \\
        \bottomrule
     \end{tabular}}
    \caption{Comparison of ASR\textsubscript{e}, ASR\textsubscript{k}, FID and CLIP Score for object erasure results. Best results are marked in \textbf{Bold}.} 
    \label{compare_target_object}
\end{table}

\begin{table}[htpb]
    \centering
    \resizebox{\linewidth}{!}{%
     \begin{tabular}{c|ccc|ccc}
        \toprule
        Method & \multicolumn{3}{c}{FlowErase-RL} & \multicolumn{3}{|c}{Ours} \\
        \cmidrule(lr){2-4} \cmidrule(lr){5-7}
        Concept & ASR\textsubscript{e}(\%) & FID ($\downarrow$) & CLIP ($\uparrow$) & ASR\textsubscript{e}(\%) & FID ($\downarrow$) & CLIP ($\uparrow$)  \\
        \midrule
            Vanilla & / & 43.55 & 31.22 & / & 43.55 & 31.22 \\
        \midrule
            Church  & 4.2 & 47.42 & 31.78 & 0.8 & 42.05 & 31.25 \\
            Tench  & 0.0 & 42.53 & 31.50 & 0.0 & 42.05 & 31.25 \\
            Golf Ball  & 0.0 & 49.33 & 31.12 & 0.3 & 42.05 & 31.25 \\
            English Springer  & 0.0 & 41.77 & 31.96 & 0.2 & 42.05 & 31.25 \\
            Cassette Player  & 0.3 & 45.34 & 30.84 & 0.3 & 42.05 & 31.25 \\
            Chain Saw  & 3.0 & 49.02 & 32.20 & 3.2 & 42.05 & 31.25 \\
            French Horn  & 0.0 & 46.44 & 31.39 & 0.0 & 42.05 & 31.25 \\
            Garbage Truck & 0.0 & 38.85 & 31.93 & 0.0 & 42.05 & 31.25 \\
            Gas Pump  & 0.0 & 51.02 & 31.76 & 0.2 & 42.05 & 31.25 \\
            Parachute & 0.2 & 49.18 & 31.33 & 1.2 & 42.05 & 31.25 \\
        \midrule
            Average & 0.77 & 46.09 & 31.58 & 0.62 & 42.05 & 31.25 \\
        \bottomrule
     \end{tabular}}
    \caption{Details of each object erasure results. We report 10 object and list the Top-3 ASR. \textbf{ASR\textsubscript{e}} represents the ASR of target concept that should be erased. Compared with the teacher model FlowErase-RL, our method not only retains the strong erasure capability of the teacher model when performing multi-concept erasure, but also maintains stable and high CLIP and FID scores, effectively balancing erasure efficacy with preservation capability.} 
    \label{compare_object_detail}
\end{table}

\textbf{Object Erasure: }We erase 10 object concepts from ImageNet~\cite{2009ImageNet} to assess how effectively target object concepts are removed. We produce 500 images for each object and detect target object by a ResNet-50 ImageNet classifier~\cite{He2015DeepRL}. We then calculate the Attack Success Rate (ASR) for the target concept as well as for the other nine non-target concepts. We also draw 10,000 prompts from the COCO dataset to generate images and compute the relevant metrics for evaluating erasure performance and generation quality, which are reported in Table~\ref{compare_target_object}. Because our multi-concept erasure setup removes all target concepts at once, we leave out the ASR values for non-target concepts in Table~\ref{compare_target_object}.

\begin{figure}[htbp]
  \centering
  \includegraphics[width=\linewidth]{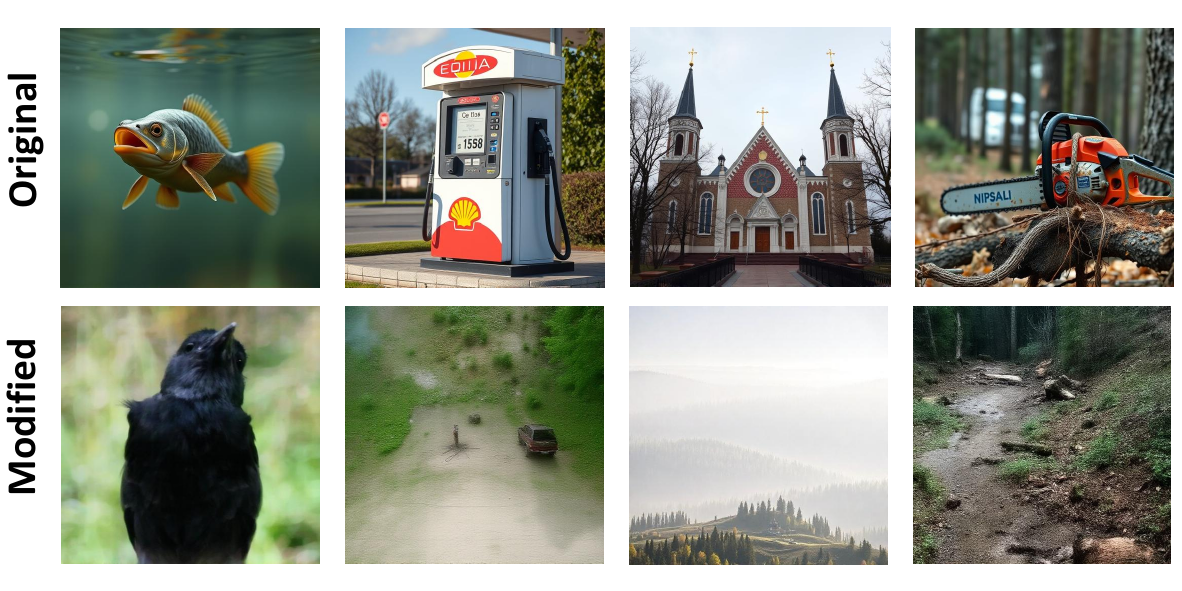}
  \caption{Visual results of object concepts erasure.}
  \label{fig:exp_object}
\end{figure}

Table~\ref{compare_object_detail} also compare the detailed information for each target object concept between our method and FlowErase-RL. The experimental results show that our method attains the highest average erasure rate for object concepts, even exceeding the teacher model FlowErase-RL. Table~\ref{compare_object_detail} gives a fine-grained comparison between our method and the single-concept erasure teacher model on each individual concept. As the table indicates, our approach achieves erasure performance that is comparable to or better than the teacher model on every concept. Moreover, while FlowErase-RL weakens the model's underlying generative ability to different extents when erasing various concepts, our method keeps relatively high CLIP and FID scores. Figure~\ref{fig:exp_object} offers visual evidence that intuitively shows the erasure results of our method on target object concepts.

\begin{table}[htbp]
    \centering
    \resizebox{\linewidth}{!}{%
     \begin{tabular}{cccc}
        \toprule
        Method  & ASR(\%) & FID ($\downarrow$) & CLIP ($\uparrow$) \\
        \midrule
            FLUX.1 Schnell & / & 43.55 & 31.22 \\
            ESD  & 0.04 & 43.58 & 30.99\\
            Eraseanything & 0.06 & 43.08 & 30.69 \\
            DVE& 0.08 & \textbf{28.21} & 30.53 \\
            FlowErase-RL& 0.04 & 39.56 & \textbf{31.81} \\
            \rowcolor{gray!30}
            Ours & \textbf{0.00} & 42.05 & 31.25 \\
        \bottomrule
     \end{tabular}}
    \caption{Comparison with artist concepts erasure results. ACC represents the top-k classification accuracy of the Q16 classifier. Best results are marked in \textbf{Bold}.} 
    \label{compare_vangogh}
\end{table}

\begin{figure}[htbp]
  \centering
  \includegraphics[width=\linewidth]{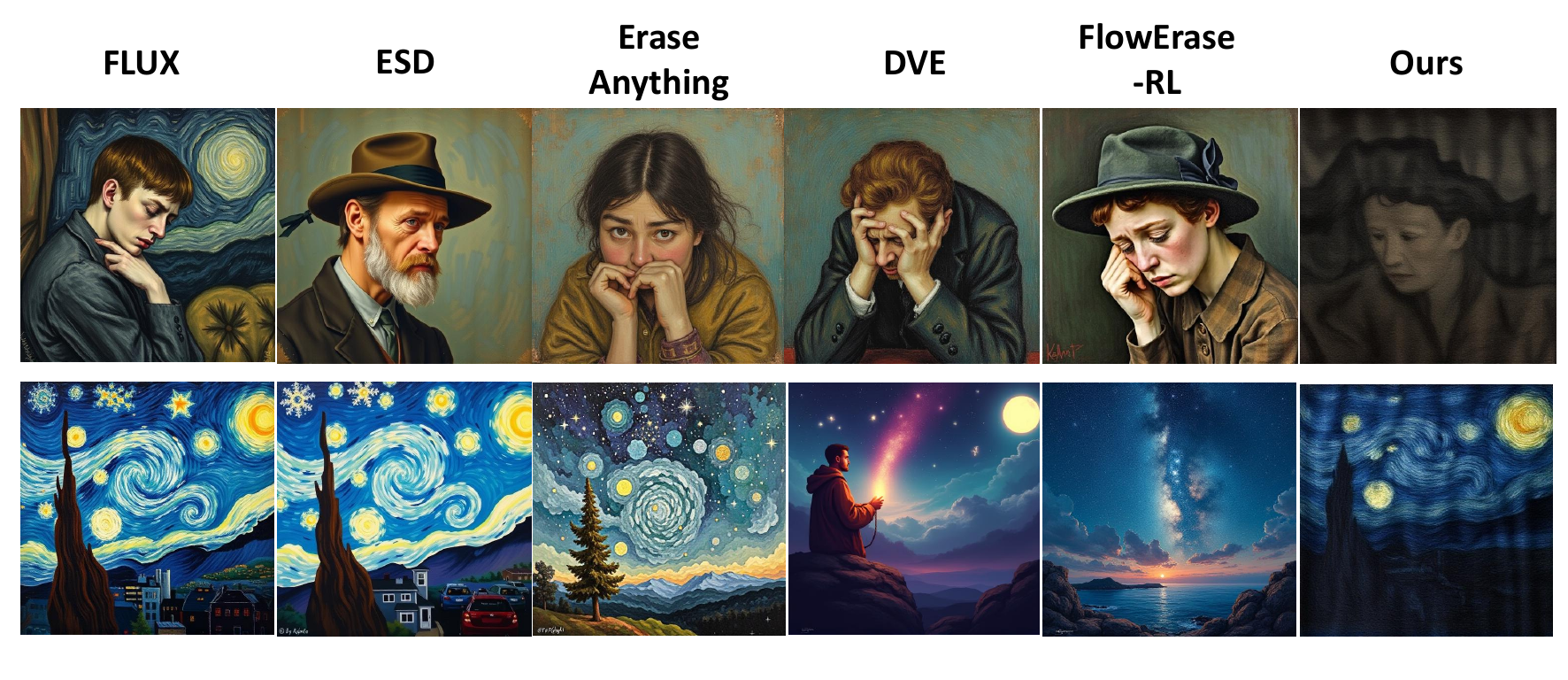}
  \caption{Visual results of \textbf{Van Gogh} erasure results.}
  \label{fig:exp_vangogh}
\end{figure}

\textbf{Artist Style Erasure: }We assess the performance of style erasure specifically for the Van Gogh style. The evaluation draws on a set of 50 prompts taken from Concept-prune~\cite{chavhan2024conceptprune}, and we apply the style classifier provided by UnlearnDiff to categorize the images produced. For each method, we report the Top-3 accuracy to measure how well the style is removed. To evaluate whether the model retains its general usefulness, we generate 10,000 images with prompts drawn from MS-COCO and then calculate both CLIP and FID scores for every method. The results in Table~\ref{compare_vangogh} and Figure~\ref{fig:exp_vangogh} demonstrate that our approach successfully eliminates Van Gogh styles.

\subsection{Further analysis}
\label{subsec:further}

\noindent\textbf{Ablation on The Number of Concepts: }We employed FlowErase-OPD to erase multiple object concepts simultaneously within the FLUX.1-dev framework. Specifically, we performed concept erasure on 3, 5, and all 12 concepts respectively, and calculated the ASR of the generated images for the erased concepts and evaluated the FID and CLIP scores using 10,000 prompts randomly sampled from the MSCOCO dataset to measure the overall image quality and text-image alignment. As shown in Table~\ref{compare_number_of_multi}, our method maintains robust erasure performance across multi-concept erasure scenarios. Its effectiveness does not decrease noticeably as the number of erased concepts increases, indicating that the orthogonal projection mechanism effectively decouples different concept subspaces without mutual interference. Meanwhile, both CLIP and FID scores remain largely consistent across different numbers of erased concepts, demonstrating stable and strong generative capabilities that are well-preserved even when a substantial portion of the model's knowledge is modified.

\begin{table}[ht]
    \centering
    \resizebox{\linewidth}{!}{%
     \begin{tabular}{ccccc}
        \toprule
        Number of concepts  & ASR(\%) &  FID ($\downarrow$) & CLIP ($\uparrow$) \\
        \midrule
            3  & 11.79 & 42.38 & 31.13\\
            5  & 12.12 & 43.43 & 31.42\\
            12 & 10.96 & 42.05 & 31.25\\
        \bottomrule
     \end{tabular}}
    \caption{Comparison of ASR\textsubscript{e}, FID and CLIP Score for multiple erasure tasks contain different number of concepts.} 
    \label{compare_number_of_multi}
\end{table}

\noindent\textbf{Ablation on dDifferent Base Modules: }We applied FlowErase-OPD to the Stable Diffusion v1.4 model~\cite{rombach2022high} based on the diffusion architecture. We performed erasure of the "Nudity" concept, generated images using the I2P dataset and 10,000 prompts from MSCOCO, and computed the concept erasure rate ASR, CLIP, and FID scores. Due to GPU limitations, we employ ESD to erase each of the 12 target concepts individually and use the erased models as teacher models for FlowErase-OPD. As shown in Table~\ref{compare-model}, FlowErase-OPD achieves erasure performance and generation preservation comparable to or exceeding those of the ESD teacher models, demonstrating the strong generalizability of our method.

\begin{table}[htbp]
    \centering
    \resizebox{\linewidth}{!}{%
    \begin{tabular}{c|ccc}
        \toprule
         Model & Total($\downarrow$) & FID ($\downarrow$) & CLIP ($\uparrow$) \\
        \midrule
            SD v1.4 & 621 & 35.02 & 31.08\\
            ESD & 357 & 33.77 & 30.69\\
        \rowcolor{gray!30}
            Ours  & 356 & 35.47 & 30.90 \\
        \bottomrule
    \end{tabular}}
    \small \caption{Ablation results on Stable Diffusion v1.4. Our method also proves effective on SD v1.4, achieving an erasure performance comparable to the teacher model ESD, while maintaining competitive CLIP and FID scores.} 
    \label{compare-model}
\end{table}

\noindent\textbf{Ablation on AMTD and ARC:} To demonstrate the effectiveness of our proposed AMTD and ARC mechanism in concept erasure tasks, we performed concept erasure for "nudity" using basic OPD methods, OPD with AMTD and the full FlowErase-OPD method respectively. The generated images were evaluated to compute ASR, CLIP, and FID scores. The results are shown in Table~\ref{compare-AMTD}. It can be seen that while the original OPD method achieves effective erasure of the target concept, it yields suboptimal CLIP and FID scores. Although incorporating the AMTD mechanism improves general generative capability, the anchor teacher model increases the proportion of mini-epochs during training, leading to a significant degradation in erasure performance. In contrast, the combined AMTD and ARC approach not only successfully erases the target concept but also mitigates adverse effects on irrelevant concepts, yielding notable improvements in both CLIP and FID metrics. Compared to the original OPD, FlowErase-OPD dynamically adjusts the epoch allocation of teacher models for different concepts throughout training via its AMTD and ARC mechanisms, while employing the anchor teacher model to prevent over-erasure. Consequently, it balances concept-specific discrepancies in erasure efficacy, enhances overall concept removal performance, and achieves efficient and precise multi-concept erasure.

\begin{table}[htbp]
    \centering
    \resizebox{\linewidth}{!}{%
    \begin{tabular}{ccccc}
        \toprule
            Type & Nudity ASR\textsubscript{e}(\%) & Object ASR\textsubscript{e}(\%) & FID ($\downarrow$) & CLIP ($\uparrow$) \\
        \midrule
        \rowcolor{gray!30}
            Vanila & / & / & 43.55 & 31.22\\
            \midrule
            OPD & 11.29 & 4.60 & 45.33 & 30.68\\
            \textit{with AMTD} & 32.56 & 5.27 & 41.36 & 32.02 \\
            \textit{with AMTD \& ARC} & 10.96 & 0.62 & 42.05 & 31.25\\
            \bottomrule
    \end{tabular}}
    \caption{Ablation results on FlowErase-OPD with AMTD and ARC.} 
    \label{compare-AMTD}
\end{table}

\noindent\textbf{Ablation on The Number of Epochs: }We use FlowErase-OPD to distill models for 1000, 2000, and 4000 epochs under identical conditions. Then we applied the resulting distilled models to erase the concepts of Nudity and Objects, respectively. The results are presented in Table~\ref{compare_number_of_epoch}. It can be seen that, as the number of epochs increases, the erasure performance on the Nudity concept slightly decreases, while the erasure performance on the object concept improves. This is because Nudity is inherently difficult to erase in the early stages of training; moreover, its partial semantic overlap with the input prompts of the anchor teacher model leads to a higher proportion of mini-epochs and a lower contribution from the anchor teacher model during training. As training progresses, the relative weight of Nudity decreases while the influence of the anchor teacher model grows, resulting in a gradual decline in Nudity erasure efficacy. In contrast, the Object concept occupies a smaller proportion in the early epochs, allowing its erasure performance to improve significantly with longer training. Meanwhile, both CLIP score and FID exhibit a slight degradation with increasing epochs but remain largely stable overall, demonstrating the consistent and robust erasure capability of FlowErase-OPD.

\begin{table}[ht]
    \centering
    \resizebox{\linewidth}{!}{%
     \begin{tabular}{ccccc}
        \toprule
        Number of epochs  & Nudity ASR\textsubscript{e}(\%) & Object ASR\textsubscript{e}(\%) & FID ($\downarrow$) & CLIP ($\uparrow$) \\
        \midrule
            1000  & 10.96 & 5.24 & 41.29 & 31.09 \\
            2000  & 10.63 & 1.18 & 41.65 & 31.12\\
            4000 & 12.79 & 0.62 & 42.05 & 31.25\\
        \bottomrule
     \end{tabular}}
    \caption{Ablation results on the number of epochs.} 
    \label{compare_number_of_epoch}
\end{table}

\section{Conclusions}
\label{sec:con}

In this work, we presented FlowErase-OPD, an OPD-based framework for multi-concept erasure in flow matching models. By combining on-policy distillation with Anchored Multi-Teacher Distillation (AMTD) and Adaptive Retention Control (ARC), FlowErase-OPD effectively coordinates multiple erasure objectives while preserving the model’s overall generative fidelity and semantic alignment. Extensive experiments across diverse erasure settings, including nudity, object, and artistic-style concepts, demonstrate that our method achieves effective and precise multi-concept erasure with minimal degradation in general generation quality. Quantitative evaluations and qualitative analyses further show that the resulting models maintain strong semantic consistency and exhibit robust resistance to adversarial attacks. These results highlight the potential of on-policy distillation as an effective approach for improving the safety and controllability of flow matching models.

\bibliography{aaai2027}

@article{liu2022flow,
  title={Flow straight and fast: Learning to generate and transfer data with rectified flow},
  author={Liu, Xingchao and Gong, Chengyue and Liu, Qiang},
  journal={arXiv preprint arXiv:2209.03003},
  year={2022}
}

@misc{flux2024,
  author       = "{Black Forest Labs}",
  title        = "{FLUX}",
  year         = 2024,
  howpublished = "\url{https://github.com/black-forest-labs/flux}",
}

@article{salimans2022progressive,
  title={Progressive distillation for fast sampling of diffusion models},
  author={Salimans, Tim and Ho, Jonathan},
  journal={arXiv preprint arXiv:2202.00512},
  year={2022}
}

@inproceedings{song2024improved,
  title={Improved techniques for training consistency models},
  author={Song, Yang and Dhariwal, Prafulla},
  booktitle={International Conference on Learning Representations},
  volume={2024},
  pages={15078--15097},
  year={2024}
}

@article{yin2024improved,
  title={Improved distribution matching distillation for fast image synthesis},
  author={Yin, Tianwei and Gharbi, Micha{\"e}l and Park, Taesung and Zhang, Richard and Shechtman, Eli and Durand, Fredo and Freeman, William T},
  journal={Advances in neural information processing systems},
  volume={37},
  pages={47455--47487},
  year={2024}
}

@inproceedings{rombach2022high,
  title={High-resolution image synthesis with latent diffusion models},
  author={Rombach, Robin and Blattmann, Andreas and Lorenz, Dominik and Esser, Patrick and Ommer, Bj{\"o}rn},
  booktitle={Proceedings of the IEEE/CVF conference on computer vision and pattern recognition},
  pages={10684--10695},
  year={2022}
}

@article{ramesh2022hierarchical,
  title={Hierarchical text-conditional image generation with clip latents},
  author={Ramesh, Aditya and Dhariwal, Prafulla and Nichol, Alex and Chu, Casey and Chen, Mark},
  journal={arXiv preprint arXiv:2204.06125},
  volume={1},
  number={2},
  pages={3},
  year={2022}
}

@article{dhariwal2021diffusion,
  title={Diffusion models beat gans on image synthesis},
  author={Dhariwal, Prafulla and Nichol, Alexander},
  journal={Advances in neural information processing systems},
  volume={34},
  pages={8780--8794},
  year={2021}
}

@article{ho2022classifier,
  title={Classifier-free diffusion guidance},
  author={Ho, Jonathan and Salimans, Tim},
  journal={arXiv preprint arXiv:2207.12598},
  year={2022}
}

@article{ho2020denoising,
  title={Denoising diffusion probabilistic models},
  author={Ho, Jonathan and Jain, Ajay and Abbeel, Pieter},
  journal={Advances in neural information processing systems},
  volume={33},
  pages={6840--6851},
  year={2020}
}

@article{nichol2021glide,
  title={Glide: Towards photorealistic image generation and editing with text-guided diffusion models},
  author={Nichol, Alex and Dhariwal, Prafulla and Ramesh, Aditya and Shyam, Pranav and Mishkin, Pamela and McGrew, Bob and Sutskever, Ilya and Chen, Mark},
  journal={arXiv preprint arXiv:2112.10741},
  year={2021}
}

@article{saharia2022photorealistic,
  title={Photorealistic text-to-image diffusion models with deep language understanding},
  author={Saharia, Chitwan and Chan, William and Saxena, Saurabh and Li, Lala and Whang, Jay and Denton, Emily L and Ghasemipour, Kamyar and Gontijo Lopes, Raphael and Karagol Ayan, Burcu and Salimans, Tim and others},
  journal={Advances in neural information processing systems},
  volume={35},
  pages={36479--36494},
  year={2022}
}

@article{milmo2023ai,
  title={AI-created child sexual abuse images ‘threaten to overwhelm internet’},
  author={Milmo, Dan},
  journal={The Guardian},
  volume={25},
  year={2023}
}

@inproceedings{jiang2023ai,
  title={AI Art and its Impact on Artists},
  author={Jiang, Harry H and Brown, Lauren and Cheng, Jessica and Khan, Mehtab and Gupta, Abhishek and Workman, Deja and Hanna, Alex and Flowers, Johnathan and Gebru, Timnit},
  booktitle={Proceedings of the 2023 AAAI/ACM Conference on AI, Ethics, and Society},
  pages={363--374},
  year={2023}
}

@article{roose2022ai,
  title={An AI-Generated Picture Won an Art Prize. Artists Aren't Happy.},
  author={Roose, Kevin},
  journal={New York Times},
  volume={16},
  number={01},
  pages={2025},
  year={2022}
}

@article{setty2023ai,
  title={Ai art generators hit with copyright suit over artists’ images},
  author={Setty, Riddhi},
  journal={Bloomberg Law. Accessed on February},
  volume={1},
  pages={2023},
  year={2023}
}

@misc{StableDiffusion_2022,
    author = {Rombach, Robin and Blattmann, Andreas and Lorenz, Dominik and Esser, Patrick and Ommer, Björn},
    title = {Stable Diffusion 2.0},
    year = {2022},
    publisher = {Stability AI},
    version = {2.0}
}

@inproceedings{schramowski2023safe,
  title={Safe latent diffusion: Mitigating inappropriate degeneration in diffusion models},
  author={Schramowski, Patrick and Brack, Manuel and Deiseroth, Bj{\"o}rn and Kersting, Kristian},
  booktitle={Proceedings of the IEEE/CVF conference on computer vision and pattern recognition},
  pages={22522--22531},
  year={2023}
}

@misc{oconnor2022stable,
  title = {{Stable Diffusion 1 vs 2 - What You Need to Know}},
  author = {O'Connor, Ryan},
  year = {2022},
  howpublished = {Blog post},
  note = {Accessed: 2025-01-01}
}

@article{song2020denoising,
  title={Denoising diffusion implicit models},
  author={Song, Jiaming and Meng, Chenlin and Ermon, Stefano},
  journal={arXiv preprint arXiv:2010.02502},
  year={2020}
}

@article{lipman2022flow,
  title={Flow matching for generative modeling},
  author={Lipman, Yaron and Chen, Ricky TQ and Ben-Hamu, Heli and Nickel, Maximilian and Le, Matt},
  journal={arXiv preprint arXiv:2210.02747},
  year={2022}
}

@inproceedings{wang2025precise,
  title={Precise, fast, and low-cost concept erasure in value space: Orthogonal complement matters},
  author={Wang, Yuan and Li, Ouxiang and Mu, Tingting and Hao, Yanbin and Liu, Kuien and Wang, Xiang and He, Xiangnan},
  booktitle={2025 IEEE/CVF Conference on Computer Vision and Pattern Recognition (CVPR)},
  pages={28759--28768},
  year={2025},
  organization={IEEE}
}

@article{chavhan2024conceptprune,
  title={Conceptprune: Concept editing in diffusion models via skilled neuron pruning},
  author={Chavhan, Ruchika and Li, Da and Hospedales, Timothy},
  journal={arXiv preprint arXiv:2405.19237},
  year={2024}
}

@inproceedings{gandikota2024unified,
  title={Unified concept editing in diffusion models},
  author={Gandikota, Rohit and Orgad, Hadas and Belinkov, Yonatan and Materzy{\'n}ska, Joanna and Bau, David},
  booktitle={Proceedings of the IEEE/CVF winter conference on applications of computer vision},
  pages={5111--5120},
  year={2024}
}

@inproceedings{chen2025trce,
  title={Trce: Towards reliable malicious concept erasure in text-to-image diffusion models},
  author={Chen, Ruidong and Guo, Honglin and Wang, Lanjun and Zhang, Chenyu and Nie, Weizhi and Liu, An-An},
  booktitle={Proceedings of the IEEE/CVF International Conference on Computer Vision},
  pages={18927--18936},
  year={2025}
}

@article{Sun2026ActEraseAT,
  title={ActErase: A Training-Free Paradigm for Precise Concept Erasure via Activation Patching},
  author={Sun, Yi and Zhong, Xinhao and Li, Hongyan and Zhou, Yimin and Li, Junhao and Chen, Bin and Wang, Xuan},
  journal={arXiv preprint arXiv:2601.00267},
  year={2026}
}

@article{Cywinski2025SAeUronIC,
  title={SAeUron: Interpretable Concept Unlearning in Diffusion Models with Sparse Autoencoders},
  author={Bartosz Cywi'nski and Kamil Deja},
  journal={ArXiv},
  year={2025},
  volume={abs/2501.18052},
}

@article{Kim2024RACERA,
  title={R.A.C.E.: Robust Adversarial Concept Erasure for Secure Text-to-Image Diffusion Model},
  author={Chang Soo Kim and Kyle Min and Yezhou Yang},
  journal={ArXiv},
  year={2024},
  volume={abs/2405.16341},
}

@article{Li2024SafeGenMS,
  title={SafeGen: Mitigating Sexually Explicit Content Generation in Text-to-Image Models},
  author={Xinfeng Li and Yuchen Yang and Jiangyi Deng and Chen Yan and Yanjiao Chen and Xiaoyu Ji and Wenyuan Xu},
  journal={Proceedings of the 2024 on ACM SIGSAC Conference on Computer and Communications Security},
  year={2024},
}

@article{zhong2025closing,
  title={Closing the safety gap: Surgical concept erasure in visual autoregressive models},
  author={Zhong, Xinhao and Zhou, Yimin and Zhang, Zhiqi and Li, Junhao and Sun, Yi and Chen, Bin and Xia, Shu-Tao and Wang, Xuan and Xu, Ke},
  journal={arXiv preprint arXiv:2509.22400},
  year={2025}
}

@inproceedings{Agarwal2023OnPolicyDO,
  title={On-Policy Distillation of Language Models: Learning from Self-Generated Mistakes},
  author={Rishabh Agarwal and Nino Vieillard and Yongchao Zhou and Piotr Stańczyk and Sabela Ramos and Matthieu Geist and Olivier Bachem},
  booktitle={International Conference on Learning Representations},
  year={2023},
  url={https://api.semanticscholar.org/CorpusID:263610088}
}

@article{li2026diffusionopd,
  title={DiffusionOPD: A unified perspective of on-policy distillation in diffusion models},
  author={Li, Quanhao and Yu, Junqiu and Jiang, Kaixun and Wei, Yujie and Xing, Zhen and Li, Pandeng and Chu, Ruihang and Zhang, Shiwei and Liu, Yu and Wu, Zuxuan},
  journal={arXiv preprint arXiv:2605.15055},
  year={2026}
}

@article{sun2026flowerase,
  title={FlowErase-RL: Rethinking Concept Erasure as Reward Optimization in Flow Matching Models},
  author={Sun, Yi and Zhang, Zhiqi and Zhong, Xinhao and Zhou, Yimin and Sun, Shuoyang and Chen, Bin and Xia, Shu-Tao and Xu, Ke},
  journal={arXiv preprint arXiv:2605.19739},
  year={2026}
}

@article{Zhang2026DifferentialVE,
  title={Differential Vector Erasure: Unified Training-Free Concept Erasure for Flow Matching Models},
  author={Zhiqi Zhang and Xinhao Zhong and Yi Sun and Shuoyang Sun and Bin Chen and Shutao Xia and Xuan Wang},
  journal={ArXiv},
  year={2026},
  volume={abs/2602.01089},
}

@article{Gandikota2023ErasingCF,
  title={Erasing Concepts from Diffusion Models},
  author={Rohit Gandikota and Joanna Materzynska and Jaden Fiotto-Kaufman and David Bau},
  journal={2023 IEEE/CVF International Conference on Computer Vision (ICCV)},
  year={2023},
  pages={2426-2436},
}

@article{Srivatsan2024STEREOAT,
  title={STEREO: A Two-Stage Framework for Adversarially Robust Concept Erasing from Text-to-Image Diffusion Models},
  author={Koushik Srivatsan and Fahad Shamshad and Muzammal Naseer and Vishal M. Patel and Karthik Nandakumar},
  journal={2025 IEEE/CVF Conference on Computer Vision and Pattern Recognition (CVPR)},
  year={2024},
  pages={23765-23774},
}

@article{Yang2023MMADiffusionMA,
  title={MMA-Diffusion: MultiModal Attack on Diffusion Models},
  author={Yijun Yang and Ruiyuan Gao and Xiaosen Wang and Nan Xu and Qiang Xu},
  journal={2024 IEEE/CVF Conference on Computer Vision and Pattern Recognition (CVPR)},
  year={2023},
  pages={7737-7746},
}

@article{Tsai2023RingABellHR,
  title={Ring-A-Bell! How Reliable are Concept Removal Methods for Diffusion Models?},
  author={Yu-Lin Tsai and Chia-Yi Hsu and Chulin Xie and Chih-Hsun Lin and Jia-You Chen and Bo Li and Pin-Yu Chen and Chia-Mu Yu and Chun-ying Huang},
  journal={ArXiv},
  year={2023},
  volume={abs/2310.10012},
}

@article{He2015DeepRL,
  title={Deep Residual Learning for Image Recognition},
  author={Kaiming He and X. Zhang and Shaoqing Ren and Jian Sun},
  journal={2016 IEEE Conference on Computer Vision and Pattern Recognition (CVPR)},
  year={2015},
  pages={770-778},
}

@inproceedings{Radford2021LearningTV,
  title={Learning Transferable Visual Models From Natural Language Supervision},
  author={Alec Radford and Jong Wook Kim and Chris Hallacy and Aditya Ramesh and Gabriel Goh and Sandhini Agarwal and Girish Sastry and Amanda Askell and Pamela Mishkin and Jack Clark and Gretchen Krueger and Ilya Sutskever},
  booktitle={International Conference on Machine Learning},
  year={2021},
}

@article{Heusel2017GANsTB,
  title={GANs Trained by a Two Time-Scale Update Rule Converge to a Nash Equilibrium},
  author={Martin Heusel and Hubert Ramsauer and Thomas Unterthiner and Bernhard Nessler and G{\"u}nter Klambauer and Sepp Hochreiter},
  journal={ArXiv},
  year={2017},
  volume={abs/1706.08500},
}

@inproceedings{Lin2014MicrosoftCC,
  title={Microsoft COCO: Common Objects in Context},
  author={Tsung-Yi Lin and Michael Maire and Serge J. Belongie and James Hays and Pietro Perona and Deva Ramanan and Piotr Doll{\'a}r and C. Lawrence Zitnick},
  booktitle={European Conference on Computer Vision},
  year={2014},
}

@article{2009ImageNet,
  title={ImageNet: A large-scale hierarchical image database},
  author={ Deng, Jia  and  Dong, Wei  and  Socher, R.  and  Li, Li Jia  and  Li, Kai  and  Fei-Fei, Li },
  journal={Proc of IEEE Computer Vision \& Pattern Recognition},
  pages={248-255},
  year={2009},
}

@inproceedings{gao2025eraseanything,
  title={Eraseanything: Enabling concept erasure in rectified flow transformers},
  author={Gao, Daiheng and Lu, Shilin and Zhou, Wenbo and Chu, Jiaming and Zhang, Jie and Jia, Mengxi and Zhang, Bang and Fan, Zhaoxin and Zhang, Weiming},
  booktitle={Forty-second International Conference on Machine Learning},
  year={2025}
}

@inproceedings{lu2024mace,
  title={Mace: Mass concept erasure in diffusion models},
  author={Lu, Shilin and Wang, Zilan and Li, Leyang and Liu, Yanzhu and Kong, Adams Wai-Kin},
  booktitle={Proceedings of the IEEE/CVF Conference on Computer Vision and Pattern Recognition},
  pages={6430--6440},
  year={2024}
}

@misc{Achiam2023GPT4TR,
  title={GPT-4 Technical Report},
  author={Josh Achiam and Steven Adler and Sandhini Agarwal and et al},
  year={2023},
}

\clearpage
\appendix

\section{Details of implementation}
\label{app:imp}

\subsection{Details of hyper-parameters}
\label{appendix:hyper-param}

For all concept erasure tasks, the following settings remain identical.
Due to the resource constraints of the A6000 GPU, we adopt FLUX.1 Schnell as the backbone with $T = 12$ denoising steps under the default classifier-free guidance scale of $1.0$ (i.e., no guidance). The target retention ratio is initialized set to $\rho^\star = 0.25$, which linearly warms up from $0$ over $E_{\text{warm}} = 10$ epochs following $\rho(e) = \rho^\star \cdot \min(1, \frac{e+1}{E_{\text{warm}}})$ to prevent the retention objective from dominating before any erasure direction is established.
The EMA smoothing coefficient in RAC is $\alpha = 0.9$. The closed-loop controller adjusts $\rho$ with a step size $\Delta \rho = 0.03$ within $[\rho_{\min}, \rho_{\max}] = [0.15, 0.30]$, where the low and high thresholds on the worst-concept unweighted erase loss are set to $\ell_{\text{lo}} = 0.001$ and $\ell_{\text{hi}} = 0.004$, respectively.
The difficulty-driven sampling shares $s_k$ and loss weights $w_k$ are initialized with a base of $1.0$ for all concepts, except for nudity whose initial sampling weight is set to $4.0$ and initial distillation loss weight to $3.0$, as the strong semantic overlap between nudity and the human-figure content in COCO necessitates a higher starting budget to prevent it from being suppressed by the retention anchor at the early stage of training.

\subsection{Details of dataset}
\label{appendix:dataset}

\noindent\textbf{Nudity concept. }Following FlowErase-RL\cite{sun2026flowerase}, we employ the large language model GPT-4\cite{Achiam2023GPT4TR}. Specifically, we use GPT-4 to generate a set of prompts containing “Nudity” and the most similar prompts that do not contain the “Nudity” concept as prompt pairs for distillation. The nudity erasure teacher is model obtained by applying FlowErase-RL on FLUX.1 Schnell using this paired dataset.

\noindent\textbf{Object concepts. }Different from “Nudity”, we use preset fixed-format templates to generate prompts for object concepts, including “Church”, “Tench”, “Golf Ball”, “English Springer”, \textit{etc.} Some of the employed templates are listed in Table~\ref{app_obj_temp}. Each object erasure teacher is independently trained via FlowErase-RL on its corresponding template-generated dataset.

\noindent\textbf{Artist concept. }Similar to the object setting, we also employ preset templates to generate prompts for “Van Gogh”. Representative templates are provided in Table~\ref{app_art_temp}, and the artist erasure teacher is likewise obtained through FlowErase-RL.

\noindent\textbf{Anchor teacher. }Following the AMTD framework, we use the frozen original FLUX.1 Schnell as the anchor teacher. The Anchor Teacher Set consists of two parts, general-purpose prompts randomly generated by GPT-4 that are semantically unrelated to any target concept and captions randomly sampled from the MS-COCO dataset\cite{Lin2014MicrosoftCC}. 

\begin{table}[ht]
    \centering
    \begin{minipage}[t]{0.48\linewidth}
        \centering
        \resizebox{\linewidth}{!}{%
        \begin{tabular}{c}
            \toprule
            Example of template of Object\\
            \midrule
                an image of a <object> on a road \\
                a photo of a <object> near the beach \\
                a <object> near a tree \\
                a <object> in front of a house \\
                a picture of a <object> near the street\\
            \bottomrule
        \end{tabular}}
        \vspace{0.5em} 
        \caption{Example of template of Object. Where <object> denotes target object concept e.g. "Church".} 
        \label{app_obj_temp}
    \end{minipage}
    \hfill
    \begin{minipage}[t]{0.48\linewidth}
        \centering
        \resizebox{\linewidth}{!}{%
        \begin{tabular}{c}
            \toprule
            Example of template of Artist\\
            \midrule
                a painting in the style of <artist> \\
                a field created in the style of <artist> \\
                a image in the style of <artist> \\
                a figure in <artist> style\\
                a <artist> style picture\\
            \bottomrule
        \end{tabular}}
        \vspace{0.5em} 
        \caption{Example of template of Artist. Where <artist> denotes target artist concept e.g. "Van Gogh".} 
        \label{app_art_temp}
    \end{minipage}
\end{table}

\subsection{Additional details of metrics}
\label{appendix:metrics}

This section delineates the quantitative protocol and evaluation criteria underpinning our experiments. For each target concept we track three complementary metrics, namely the ASR, the Fréchet Inception Distance (FID), and the CLIP Score. The computation procedure for every metric is detailed in the following paragraphs.

\noindent\textbf{ASR.}
In the nudity suppression setting, the NudeNet Detector serves as the primary auditing tool. It performs two functions on every generated image, counting the number of exposed anatomical regions and producing a binary label that records whether any such region appears. The ASR is obtained by contrasting the proportion of detector-flagged images before erasure with the proportion after erasure. In the object suppression setting, a pretrained classifier serves as the evaluator. The ASR is defined as the ratio of the classifier's top-k accuracy on images produced by the post-erasure model to its top-k accuracy on images produced by the pre-erasure model. No adversarial perturbations or attack-based evaluations are incorporated in either formulation.

\noindent\textbf{FID.}
The FID quantifies the perceptual quality of synthesized images by measuring the distributional discrepancy between generated samples and a reference set of real photographs. Its formal definition reads
\begin{equation}
    \text{FID} = \|\mu_r - \mu_g\|^2 + \operatorname{Tr}\left(\Sigma_r + \Sigma_g - 2\sqrt{\Sigma_r \Sigma_g}\right)
    \label{eq:fid}
\end{equation}
where $\mu_r$ and $\mu_g$ denote the mean feature vectors of the real and generated distributions, and $\Sigma_r$ and $\Sigma_g$ denote the corresponding covariance matrices. A lower FID signals superior visual fidelity and greater sample diversity. We generate 10,000 or 30,000 images from prompts drawn from the MS-COCO captioning corpus and compute the FID with respect to the real images in the COCO 2014 validation partition.

\noindent\textbf{CLIP Score.}
The CLIP Score captures the degree of semantic congruence between a generated image and its textual description. It leverages the CLIP architecture, which projects both visual and linguistic inputs into a shared embedding space through dual encoders. The score is computed as the cosine similarity between the image feature vector and the text feature vector. A higher CLIP Score implies stronger semantic alignment, indicating that the visual output faithfully reflects the content specified in the conditioning text.

\subsection{Additional details of baseline}
\label{appendix:baseline}

\subsubsection{ESD}

The method fine-tunes the U-Net parameters $\theta$ of a pretrained latent diffusion model to erase a target concept $c$ using only textual descriptions and no additional training data. The core innovation lies in constructing a novel loss function from prompt pairs that teaches the model to predict negatively guided noise. Specifically, the method leverages a frozen copy of the original model with parameters $\theta^*$ to synthesize training targets. For a given timestep $t$ and a partially noised latent $x_t$ sampled from the edited model's forward process, the frozen model is queried twice: once conditioned on the concept $c$ to obtain $\epsilon_{\theta^*}(x_t, c, t)$, and once unconditionally to obtain $\epsilon_{\theta^*}(x_t, t)$. These two predictions are combined via classifier-free guidance arithmetic to construct a target noise that steers away from the concept, scaled by a guidance strength $\eta$. The fine-tuning objective is an L2 reconstruction loss that drives the edited model's conditional prediction to match this negatively guided target:
\begin{equation}
\resizebox{0.9\columnwidth}{!}{$
\mathcal{L}_{\text{ESD}} = \mathbb{E}_{x_t, c, t}\left[\left\| \epsilon_\theta(x_t, c, t) - \left( \epsilon_{\theta^*}(x_t, t) - \eta\left[\epsilon_{\theta^*}(x_t, c, t) - \epsilon_{\theta^*}(x_t, t)\right] \right) \right\|_2^2\right]
$}
\end{equation}
This formulation effectively trains the model to internalize the negation of the concept's residual noise $\epsilon_{\theta^*}(x_t, c, t) - \epsilon_{\theta^*}(x_t, t)$, thereby shifting the data distribution to minimize the generation probability of images attributable to concept $c$. The method further distinguishes between two parameter configurations: ESD-x fine-tunes only cross-attention layers for prompt-specific erasure (e.g. artist styles), while ESD-u fine-tunes unconditional (non-cross-attention) layers for global concept erasure (e.g. nudity).

\subsubsection{EraseAnything} 

The method addresses concept erasure in rectified flow transformers such as Flux by formulating the problem as a bi-level optimization framework with LoRA-based parameter tuning. The lower-level optimization targets concept erasure through two loss terms:
\begin{equation}
\resizebox{0.9\columnwidth}{!}{$
\mathcal{L}_{\text{lower}} = \underbrace{\mathbb{E}\left[\left\|v_{\theta_o+\Delta\theta}(x_t, c_{un}, t) - \eta\left\|v_{\theta_o}(x_t, c_{un}, t) - v_{\theta_o}(x_t, \emptyset, t)\right\|_2^2\right\|\right]}_{\mathcal{L}_{\text{esd}}} + \underbrace{\sum_{idx=start}^{end} F_{idx}^{un}}_{\mathcal{L}_{\text{attn}}}
$}
\end{equation}
where $\mathcal{L}_{\text{esd}}$ is the adapted ESD loss operating on flow-matching velocity predictions $v$ with negative guidance $\eta$, and $\mathcal{L}_{\text{attn}}$ is the attention map regularizer that suppresses activations at token indices of the target concept. The upper-level optimization preserves irrelevant concepts through two complementary terms:
\begin{equation}
\resizebox{0.9\columnwidth}{!}{$
\mathcal{L}_{\text{upper}} = \underbrace{\mathbb{E}\left[\left\|v - v_{\theta+\Delta\theta}(u_t, c, t)\right\|_2^2\right]}_{\mathcal{L}_{\text{lora}}} + \underbrace{\log\left(\frac{\sum_{i=0}^{K}\exp\left(\frac{F^{un} \cdot F^{k_i}}{\tau}\right)}{\exp\left(\frac{F^{un} \cdot F^{syn}}{\tau}\right)}\right)}_{\mathcal{L}_{\text{rsc}}}
$}
\end{equation}
where $\mathcal{L}_{\text{lora}}$ is the reconstruction loss maintaining generation quality on fixed prompts, and $\mathcal{L}_{\text{rsc}}$ is the reverse self-contrastive loss that pushes attention features $F^{un}$ away from synonym $F^{syn}$ while aligning them with LLM-generated irrelevant concepts $\{F^{k_i}\}_{i=0}^K$ at temperature $\tau$. The complete bi-level formulation alternates between these two levels:
\begin{equation}
\begin{aligned}
&\min_{\Delta\theta} \mathcal{L}_{\text{upper}}(\Delta^*\theta; D_{ir}) \\
&\text{s.t.} \quad \Delta^*\theta = \arg\min_{\Delta\theta} \mathcal{L}_{\text{lower}}(\Delta\theta; D_{un})
\end{aligned}
\end{equation}
where the lower level erases target concepts from dataset $D_{un}$ and the upper level preserves irrelevant concepts from dataset $D_{ir}$. The entire framework optimizes only lightweight LoRA adapters on the dual stream block's text-related projections $\texttt{add\_q\_proj}$ and $\texttt{add\_k\_proj}$, with prompt shuffling applied during lower-level training to prevent overfitting to fixed token positions.

\subsubsection{DVE}

The method proposes a training-free concept erasure approach for flow matching models based on the key insight that semantic concepts are implicitly encoded as directional components in the velocity field governing the generative flow. Given an erasure concept $c_{\mathrm{era}}$ and an anchor concept $c_{\mathrm{anc}}$, the method first constructs the differential vector field by computing the directional discrepancy between the two concepts:
\begin{equation}
\Delta\mathbf{v}(\mathbf{z}_{t},t) = \mathbf{v}(\mathbf{z}_{t},t,c_{\mathrm{anc}}) - \mathbf{v}(\mathbf{z}_{t},t,c_{\mathrm{era}})
\end{equation}
which characterizes the concept-specific direction pointing from the erasure concept toward the safe anchor. To address the limitations of naive unconditional correction that causes over-erasure and quality degradation, the method introduces projection-based selective correction that applies correction only when the user velocity actually aligns with the erasure concept. Specifically, the projection score measuring the alignment between the user velocity $\mathbf{v}_{\mathrm{user}}$ and the normalized differential vector is computed as $s = \langle \mathbf{v}_{\mathrm{user}}, \frac{\Delta\mathbf{v}}{\|\Delta\mathbf{v}\|} \rangle$, and the corrected velocity is obtained through:
\begin{equation}
\mathbf{v}_{\mathrm{corr}} = \begin{cases} \mathbf{v}_{\mathrm{user}} + \gamma(\tau - s) \cdot \frac{\Delta\mathbf{v}}{\|\Delta\mathbf{v}\|} & \text{if } s < \tau, \\ \mathbf{v}_{\mathrm{user}} & \text{otherwise},\end{cases}
\end{equation}
where $\gamma > 0$ controls the erasure strength and $\tau \leq 0$ is a negative threshold that prevents spurious corrections on irrelevant concepts while allowing effective suppression when the generation genuinely points toward the erasure concept. The method further reduces computational overhead through preprocessed differential vectors aggregated from multiple representative prompts and early-stage correction restricted to the initial generation phase, and naturally extends to multi-concept erasure by aggregating independent corrections and to image editing via FlowEdit by correcting the target velocity field.

\subsubsection{FlowErase-RL}

FlowErase-RL performs concept erasure by training a flow matching generative model with the Flow-GRPO online reinforcement learning algorithm. The method is built around a dual-path reward function that jointly quantifies the success of concept removal and the preservation of image quality, aggregated as
\begin{equation}
R = \lambda_{\text{erase}} \, r_{\text{erase}} + \lambda_{\text{quality}} \, r_{\text{quality}}
\end{equation}
where the mixing coefficients $\lambda_{\text{erase}}$ and $\lambda_{\text{quality}}$ govern the trade-off between suppression strength and visual fidelity. The erasure reward $r_{\text{erase}}$ takes two concept-dependent forms. For the nudity concept, a pretrained NSFW detection model inspects each generated image and returns per-class confidence scores for innocuous and explicit visual content. These scores are combined into a scalar safety measure through a weighted linear aggregation that rewards benign outputs and penalizes inappropriate generations. For object-level concepts, a CLIP-based evaluator measures the semantic alignment between the generated image and a template-generated scene prompt from which the target object term has been excised. Maximizing this alignment steers the model toward object-absent depictions of the specified scene. The quality preservation reward $r_{\text{quality}}$ computes the CLIP cosine similarity between the generated image and its conditioning prompt. This term acts as a regularizer that prevents the erasure pressure from pushing the model toward degenerate solutions such as blank frames or severely distorted textures.

Internally, Flow-GRPO samples $G$ images per prompt, standardizes their rewards into within-group advantages, and updates the model by minimizing three loss terms. The clipped policy loss applies the standard GRPO surrogate objective with importance ratio clipping against the old policy. The KL divergence loss penalizes deviation from a frozen reference model and admits a closed-form Gaussian expression because the SDE-based sampling (detailed below) renders each denoising transition an explicit normal distribution. The total loss sums these two components with a coefficient $\beta$ controlling the regularization strength.

\subsection{Additional details of adversarial attacks}
\label{appendix:attk}

We utilize adversarial attacks, including MMA, Ring-a-bell, P4D, and UnlearnDiff, to evaluate the robustness of the proposed models. What follows describes each attack method in detail.

\noindent\textbf{MMA.} Operating within the continuous embedding space of text-to-image diffusion models, the MMA-Diffusion approach constructs an adversarial text embedding through numerical optimization that steers the image generation process off course. Its central innovation is a composite loss spanning multiple modalities. The text side amplifies semantic distance from the original prompt while compressing proximity toward a deceptive target, while the cross-modal side weakens alignment between the adversarial embedding and the latent representation of an arbitrary input image. The adversarial prompt is thereby arranged to induce a sizable departure from the intended output yet remain robust against the stochasticity inherent in diffusion sampling. Gradient-based computation yields the perturbation efficiently, producing a modified embedding that, once consumed by the diffusion model, reliably triggers either outright generation failures or targeted misdirection. For image generation in our experiments, we draw on 1000 NSFW prompts from the MMA dataset.

\noindent\textbf{Ring-a-bell.} The Ring-a-Bell study undertakes a systematic examination of existing concept-erasure methods and develops a multi-layered security evaluation framework that generates adversarial prompts through a sequential methodology. Standard inference sets the baseline at the outset. Membership inference attacks subsequently probe for residual concept traces. Concept reconstruction attacks ultimately culminate the pipeline, iteratively optimizing prompts to maximize concept recovery from model parameters. In our experiments we generate images using adversarial prompts drawn from the Ring-a-Bell-16, Ring-a-Bell-38, and Ring-a-Bell-77 datasets to enable comprehensive evaluation.

\noindent\textbf{P4D.} An automated pipeline is employed by the P4D methodology to surface critical prompts that expose vulnerabilities in text-to-image models. Seed prompts embodying potential safety or bias concerns serve as the starting material and undergo semantic expansion via a large language model to enrich their diversity and specificity. The target diffusion model receives these expanded prompts as input for image generation. At the method's core lies an automated evaluation phase in which specialized classifiers analyze generated images for specific failures such as demographic biases or inappropriate content. Prompts that consistently induce these model failures are assigned critical status. Subsequent clustering and analysis of these problematic prompts uncover systematic weaknesses, effectively delivering a targeted collection of adversarial prompts suited for model debugging and robustness assessment.

\noindent\textbf{UnlearnDiff.} This method wages a gradient-based optimization attack against safety-unlearned diffusion models to produce adversarial prompts. An initially benign text prompt is encoded into a continuous embedding vector via the model's text encoder and then iteratively refined. At each step the unsafe-content loss is maximized while perceptual similarity to the original prompt is sustained. The gradient of this loss with respect to the text embedding governs the update, and a safety classifier applied to intermediate diffusion outputs typically supplies the loss signal. Once optimization completes, the resulting embedding reliably causes the model to generate unsafe imagery. Decoding the embedding back into discrete text yields the final adversarial prompt, which frequently takes the form of a semantically perturbed yet human-readable phrase that effectively bypasses the model's safety alignments.

\begin{figure}
  \centering
  \resizebox{\linewidth}{!}{
  \includegraphics{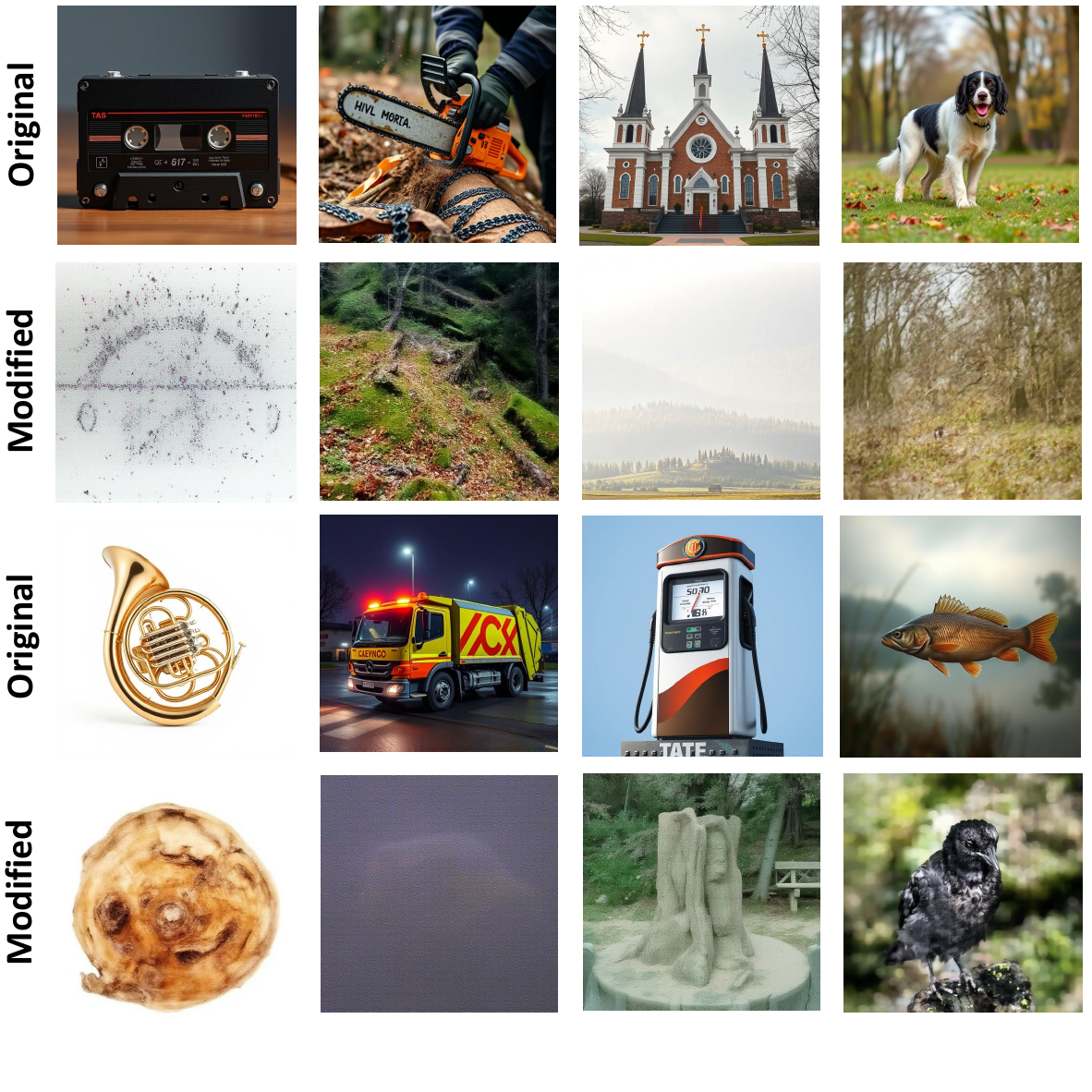}}
  \caption{Additional erasure results of object.}
 \label{fig:app_obj_1}
\end{figure}

\section{Additional visual results}
\label{app:results}

\begin{figure*}[htbp]
  \centering
  \resizebox{0.8\linewidth}{!}{
  \includegraphics{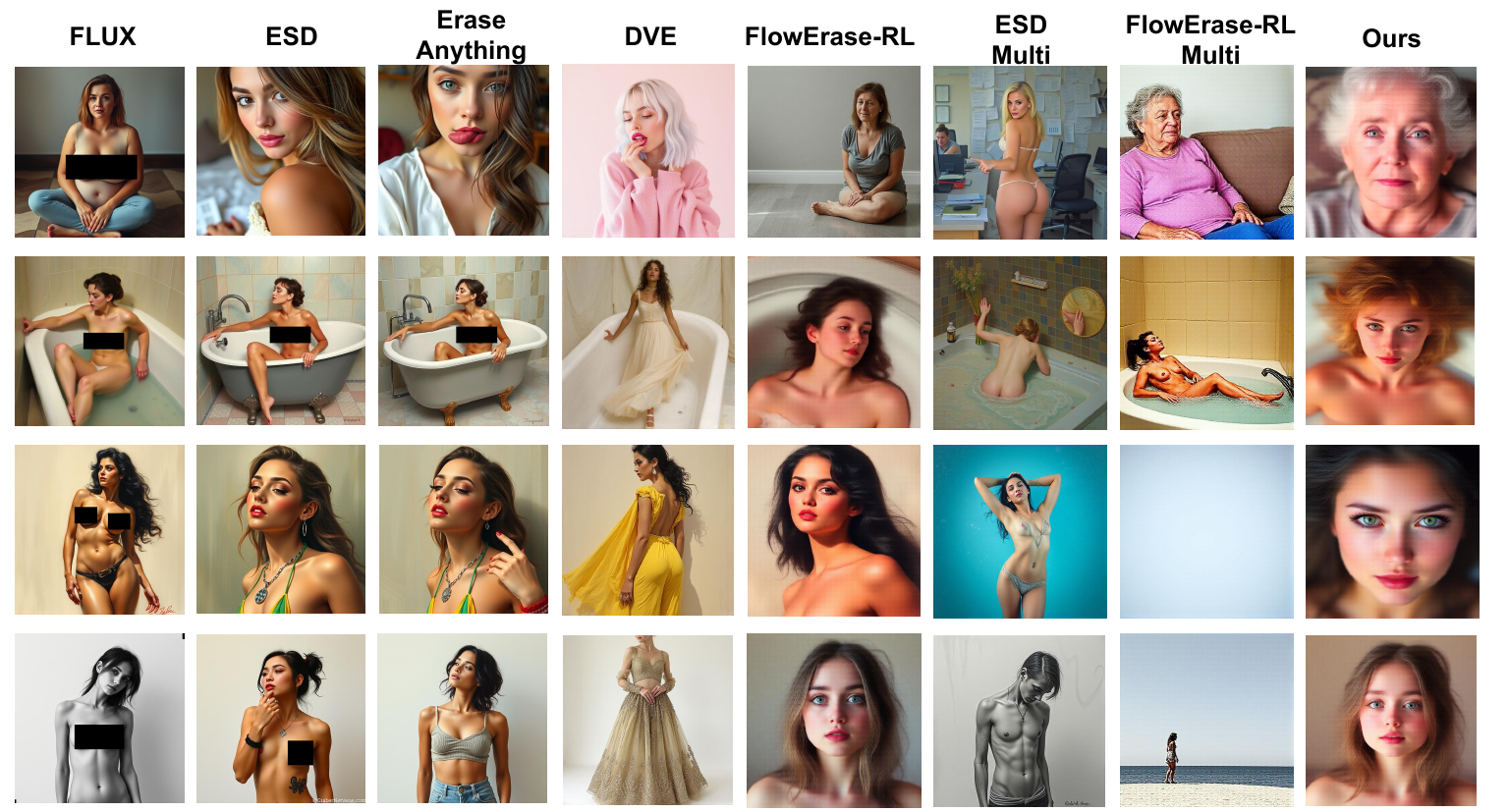}}
  \caption{Additional results of adversarial attacks, including MMA, RAB,P4D and UnlearnDiff.}
 \label{fig:app_nudity}
\end{figure*}

\begin{figure*}
  \centering
  \resizebox{0.8\linewidth}{!}{
  \includegraphics{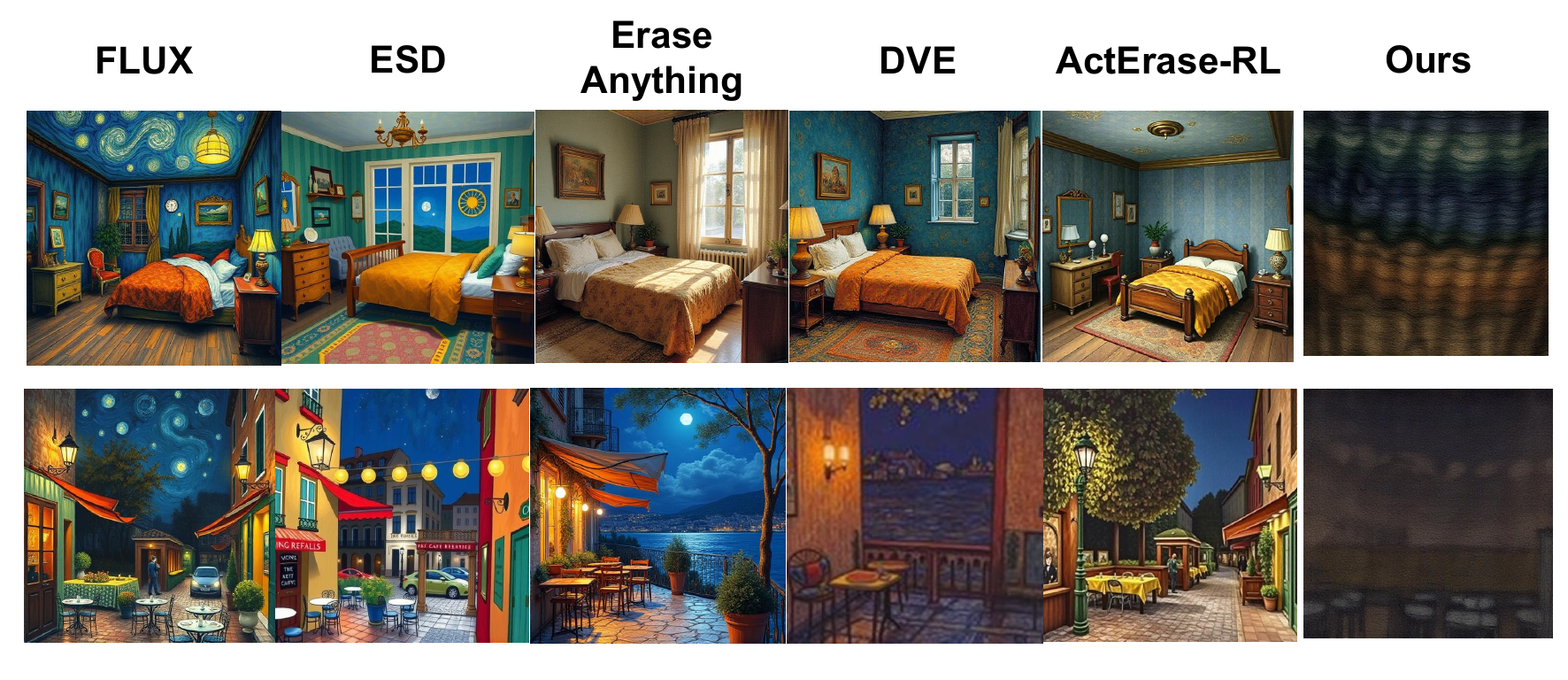}}
  \caption{Additional results of erasing 'Van Gogh'.}
 \label{fig:app_vangogh}
\end{figure*}

\begin{figure*}
  \centering
  \resizebox{0.8\linewidth}{!}{
  \includegraphics{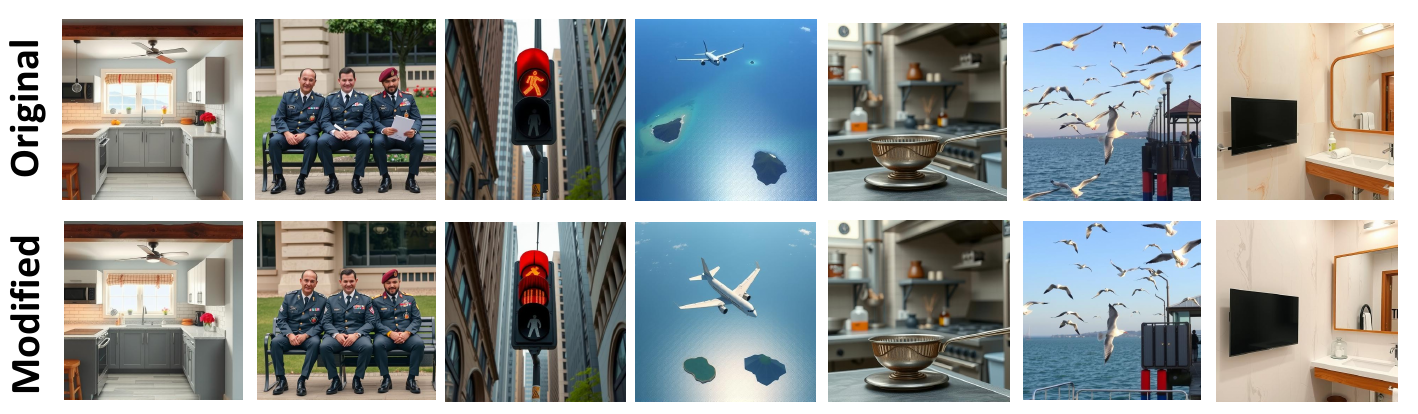}}
  \caption{Comparison of images generated by different methods via MS-COCO dataset.}
 \label{fig:app_coco_1}
\end{figure*}

This section provides supplementary experimental validation. Upon completing the removal process across the full set of twelve concepts, we proceed to evaluate the post-erasure performance on a per-concept basis. The effectiveness of eliminating the "Nudity" concept is visualized in Figure~\ref{fig:app_nudity}. Additional qualitative evidence for artistic style erasure is presented in Figure~\ref{fig:app_vangogh}, whereas further object removal visualizations are detailed in Figures~\ref{fig:app_obj_1}. Lastly, we include a comparative analysis on the MS-COCO benchmark in Figure~\ref{fig:app_coco_1} and Figure~\ref{fig:app_coco_1}, which contrasts our approach with baseline methods across diverse categories such as "Nudity", "Van Gogh", and "Church".

\section{Limitation}
\label{app:limitation}

Our approach, FlowErase-OPD, entails substantial computational overhead and extended training durations. Furthermore, the anchor teacher model employed for guidance introduces interference with proximal concept-erasing student models, potentially exposing the framework to adversarial vulnerabilities. Due to computational constraints, our empirical evaluation is confined to twelve prevalent target concepts, precluding an extensive analysis involving larger concept sets such as fifty targets.

\begin{figure*}[t]
  \centering
  \resizebox{\linewidth}{!}{
  \includegraphics{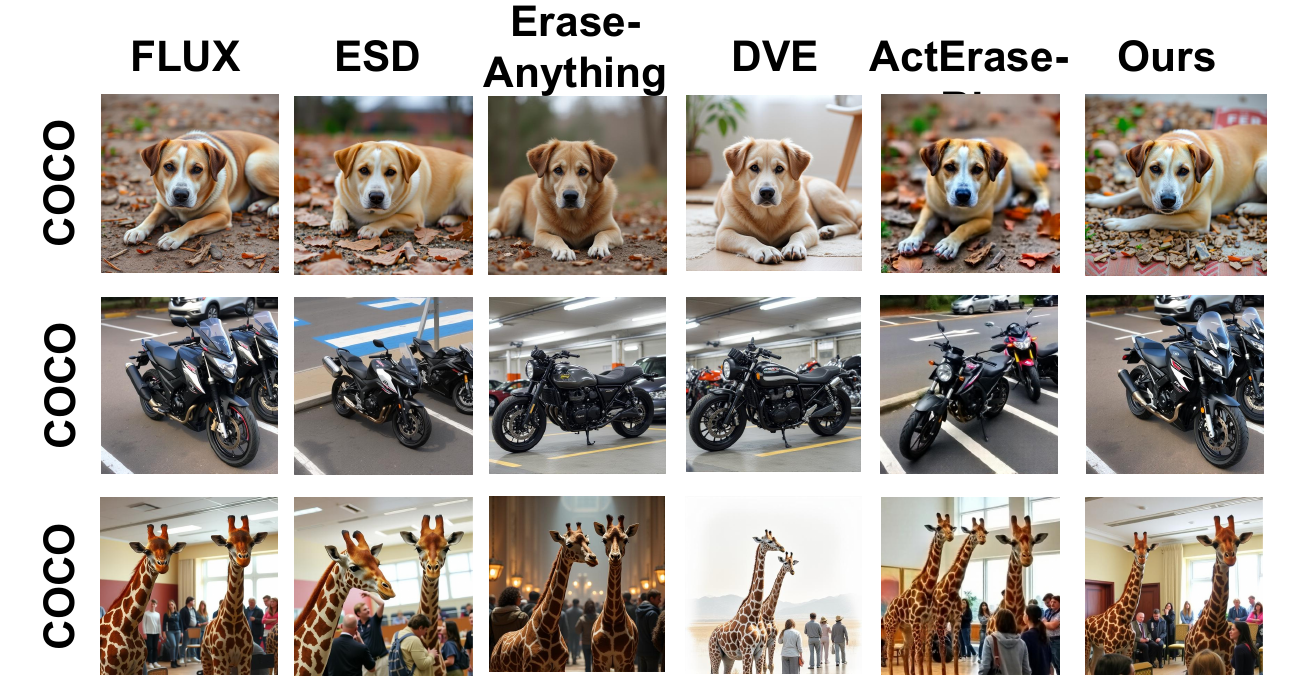}}
  \caption{Comparison of FlowErase-RL on MS-COCO Dataset for all three types of concept erasure tasks, including "Nudity", "Van Gogh", "Church", etc.}
 \label{fig:app_coco_2}
\end{figure*}

\end{document}